\documentclass{article}

\usepackage[preprint]{neurips_2026}

\usepackage[utf8]{inputenc} 
\usepackage[T1]{fontenc}    
\usepackage{hyperref}       
\usepackage{url}            
\usepackage{booktabs}       
\usepackage{amsfonts}       
\usepackage{nicefrac}       
\usepackage{microtype}      
\usepackage{xcolor}         

\usepackage{pgfplots}
\pgfplotsset{compat=1.18}

\usepackage{booktabs}
\usepackage{multirow}
\usepackage{amssymb} 
\usepackage{amsmath}
\usepackage[table,xcdraw]{xcolor}
\usepackage{scalefnt}
\usepackage{enumitem}

\title{Reviving the Discarded High-Resolution Feature for Transformer-Based Tiny Object Detection}

\author{
  Zehao Wang$^{1,2}$ \quad Si Wu$^{1,2}$ \quad Guiping Cao$^{2*}$ \quad Xin Li$^{2*}$ \quad Yaowei Wang$^{3,2}$\\
  $^1$South China University of Technology, Guangzhou, China \\
  $^2$Pengcheng Laboratory, Shenzhen, China \\
  $^3$Harbin Institute of Technology, Shenzhen, China \\
  \texttt{wangzh07@pcl.ac.cn}, \texttt{cswusi@scut.edu.cn}, \texttt{caogp@pcl.ac.cn}, \\
  \texttt{xinlihitsz@gmail.com}, \texttt{wangyaowei@hit.edu.cn}
}

\begin{document}

\maketitle

\begingroup
  \renewcommand{\thefootnote}{}
  \footnotetext{$^*$ Corresponding author.}
\endgroup

\begin{abstract}
    Tiny object detection remains challenging because tiny instances occupy only a few pixels, and their fine-grained details are easily lost during feature downsampling. Although the highest-resolution feature preserves rich local details for tiny objects, DETR-like detectors often discard it due to its large computational cost, while directly reintroducing this feature also brings dense background noise. To address this problem, we propose \textbf{ReDHF}, a novel framework that \textbf{Re}vives the \textbf{D}iscarded \textbf{H}igh-resolution \textbf{F}eature for tiny object detection. ReDHF has three cooperative components. Cross-Scale Deformable Fusion (CSDF) first constructs a semantically enhanced highest-resolution feature and injects its fine-grained details into the shallow feature under a moderate cost. Ellipse-Guided Shallow Supervision (EGSS) then guides the enhanced feature to learn more reliable classification and localization priors by applying geometry-adaptive auxiliary supervision. Full-Scale Query Interaction (FSQI) further enables decoder queries to access both low-level details and high-level semantic information, allowing them to aggregate richer cues for tiny objects. Extensive experiments show that ReDHF consistently improves multiple baselines across different DETR variants and achieves state-of-the-art performance on the evaluated benchmarks. The best ReDHF variants surpass their baselines by \textbf{+1.4 AP} on AI-TOD-v2, \textbf{+1.6 AP} on VisDrone, and \textbf{+1.5 AP} on SODA-D, with especially notable gains on tiny and small objects.

\end{abstract}



\section{Introduction}
\begin{figure}[t] 
  \centering
  \includegraphics[width=1.0\linewidth]{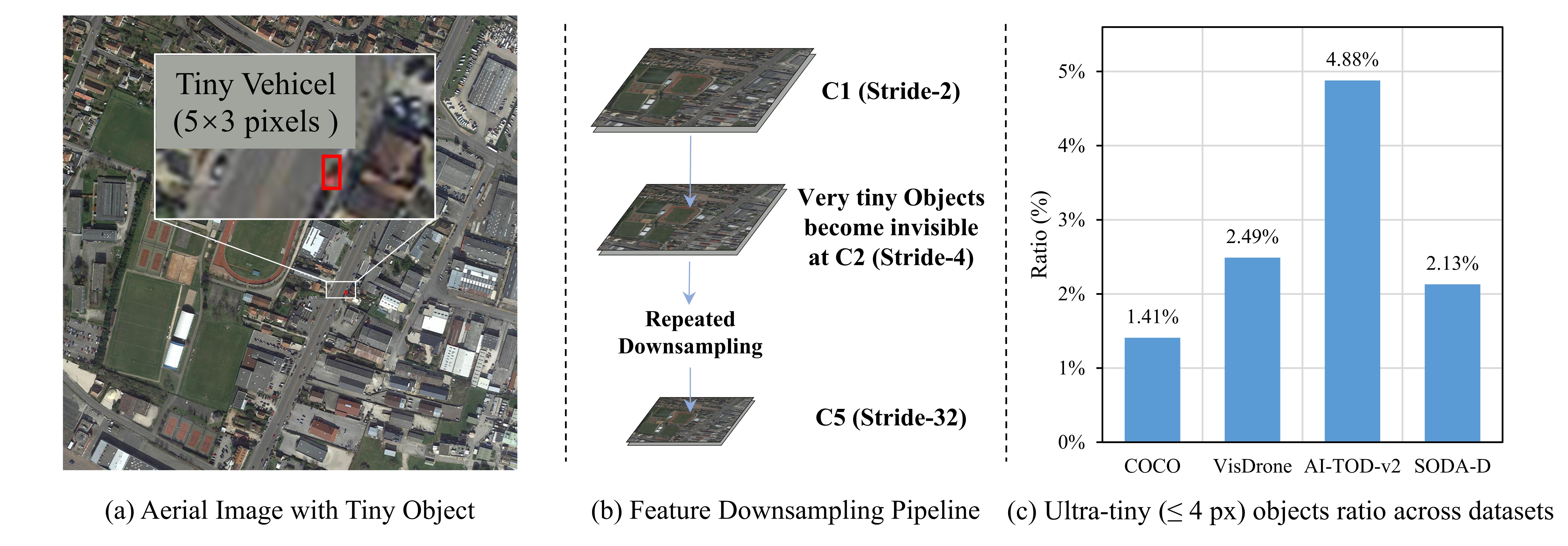}
  
  \caption{(a) Tiny objects may occupy only a few pixels, making their visual evidence extremely weak. (b) Very tiny objects can become nearly invisible on stride-4 feature map in the feature downsampling pipeline. (c) The ratio of ultra-tiny objects (short side $\leq 4$ pixels) is much higher in tiny object datasets such as AI-TOD-v2, VisDrone, and SODA-D than in COCO.}
  
  \label{fig:introduction}
\end{figure}

Object detection has advanced rapidly, with both CNN-based detectors~\cite{dai2016r-fcn, girshick2015fast, he2017mask, FPN, liu2016ssd, peng2024scene, redmon2016yolo, ren2015faster, tang2024hic, wang2023gold} and Transformer-based detectors~\cite{carion2020end, chen2024lw, huang2024deim, liu2022dabdetr, meng2021conditional, peng2024dfine, zhang2022dino, zhao2024detrs, zhu2020deformable} achieving strong performance. However, detecting tiny objects remains difficult, especially in aerial, remote sensing, and traffic scenarios. Tiny objects occupy only a few pixels, leaving their edges, textures, and geometric cues weak in the image and further weakened by feature downsampling. As a result, high-level features often fail to preserve sufficient discriminative details for tiny objects, as illustrated in Fig.~\ref{fig:introduction}(b).

To address the issues of tiny object detection, a natural idea is to preserve these weak cues by using higher-resolution inputs or features. Prior works have shown that high-resolution information is important for tiny objects. For example, Deformable DETR~\cite{zhu2020deformable} points out that small objects rely on high-resolution feature maps, and DM-EFS~\cite{sharma2025dm} shows that shallower high-resolution features can improve tiny-object representation. However, many detectors~\cite{liu2018path, zhang2022dino, zhu2020deformable, liu2022dabdetr} still discard the highest-resolution backbone feature to control computation. This design saves overhead but sacrifices the local details that are critical for tiny objects. As shown in Fig.~\ref{fig:introduction}(a)(c), discarding this feature becomes especially detrimental on AI-TOD-v2~\cite{Aitod-v2}, VisDrone~\cite{zhu2021visdrone}, and SODA-D~\cite{SODA}, where ultra-tiny objects with a short side no larger than 4 pixels appear frequently and may be compressed to nearly a single point even on a stride-4 feature map. Meanwhile, directly reintroducing the highest-resolution backbone feature is also impractical, \emph{since it has a much larger spatial size that may lead to unacceptable extra cost, and its shallow responses contain dense background activations that may overwhelm weak tiny-object signals.} These issues highlight the pressing need for detection frameworks that can balance fine-grained information preservation with computational efficiency. 


In this work, we propose \textbf{ReDHF}, a novel and efficient plug-and-play high-resolution detail reintroduction framework for Deformable-DETR-like detectors, which revives the discarded highest-resolution backbone feature while controlling computation and background disturbance through three cooperative components.
First, to address the contradiction between the limited representation of small objects and the high cost of utilizing high-resolution features, we design a lightweight \textbf{Cross-Scale Deformable Fusion (CSDF)} module that semantically enhances the highest-resolution feature to provide more useful fine-grained cues with moderate cost. Second, considering that high-resolution features mainly include low-level edges and textures and are more likely to introduce background noise, we apply \textbf{Ellipse-Guided Shallow Supervision (EGSS)} to guide the shallow feature to learn more stable classification and localization priors. Moreover, since final predictions in DETR-like detectors are query-driven, we further propose \textbf{Full-Scale Query Interaction (FSQI)}, which integrates the semantically enhanced highest-resolution feature together with higher-level features into the encoder memory, so that decoder queries can directly access pixel-level details and improve their ability of detecting tiny objects. Overall, ReDHF makes more effective use of the highest-resolution backbone feature with only a modest extra cost and significantly improves tiny object detection.

Extensive experiments show that ReDHF consistently improves multiple DETR-like~\cite{zhang2022dino, zhu2020deformable, dome-detr} baselines and achieves state-of-the-art performance. Specifically, the best ReDHF model achieves 35.7\% AP on AI-TOD-v2~\cite{Aitod-v2}, improving AP$_{vt}$ to 20.8\% (+1.2\%) over its corresponding baseline. In addition, the ReDHF variant built upon DINO-4scale~\cite{zhang2022dino} reduces GFLOPs by 86.7\% compared with directly integrating the highest-resolution feature, as detailed in Appendix~\ref{app:efficiency_analysis}. On VisDrone~\cite{zhu2021visdrone} and SODA-D~\cite{SODA}, ReDHF again brings consistent gains. The best ReDHF model achieves 40.6\% AP and 38.7\% AP, and also improves AP$_{s}$ to 33.1\% (+1.6\%) and AP$_{eS}$ to 21.4\% (+1.7\%), respectively. These results support our claim that the highest-resolution feature, though often discarded for efficiency, can be effectively revived through semantic enhancement. In summary, our main contributions are:

\begin{itemize}[itemsep=2pt]
  \item We revisit the commonly discarded highest-resolution backbone feature in DETR-like detectors and show that its effective reuse requires addressing not only computational cost, but also weak semantics and noisy shallow responses.
  
  \item We propose ReDHF, which revives the discarded feature through semantically enhanced highest-resolution feature construction (CSDF), training-time shallow auxiliary supervision (EGSS), and decoder full-scale query interaction (FSQI), thereby enabling efficient high-resolution detail reintroduction for tiny-object detection.
 \item We validate ReDHF on multiple DETR-like baselines across AI-TOD-v2, VisDrone, and SODA-D. It achieves state-of-the-art performance and consistently improves the baselines with average gains of +1.5 AP$_{vt}$, +1.7 AP$_s$, and +1.8 AP$_{eS}$ on the three datasets, respectively.
\end{itemize}

\section{Related work}
\subsection{Tiny object detection}
Identifying tiny objects is inherently difficult because tiny instances provide limited visual evidence and are vulnerable to complex background noise. Early CNN-based detectors, such as Faster R-CNN~\cite{girshick2015fast, ren2015faster}, often struggle with such objects due to inadequate feature representation and limited context modeling. Some works improve tiny-object learning by enriching training samples, such as copy-pasting augmentation~\cite{kisantal2019augmentation} and instance-level transformations~\cite{zoph2020learning}. Recently, transformer-based models~\cite{carion2020end, li2022dn, liu2022dabdetr, peng2024dfine, huang2024deim, zhang2022dino, zhao2024detrs, zhu2020deformable} have shown strong potential by bypassing hand-crafted post-processing like NMS and leveraging self-attention mechanisms. Cross-DINO~\cite{cao2025cross} combines deep MLP representations with transformer encoder features through a Cross Coding Twice Module to enhance small-object details. DQ-DETR~\cite{huang2024dq} introduces density maps to adjust query numbers and positions based on instance density, while Dome-DETR~\cite{dome-detr} further improves tiny object detection by generating density-oriented features. However, these methods often rely heavily on specific density modeling or carefully designed query mechanisms, which may complicate the optimization process and limit their flexibility across different datasets.

\subsection{Multi-scale and high-resolution features}

Feature pyramids~\cite{FPN} are widely used to handle object scale variation. Advanced architectures like BiFPN~\cite{bifpn} and DetectoRS~\cite{detectors}, improve cross-level information flow through bidirectional fusion or recursive feedback. For tiny objects, fusion factors~\cite{fusion} further regulate adjacent pyramid levels to reduce the learning burden. Recent works also explore high-resolution information more explicitly. QueryDet~\cite{querydet} refines sparse locations from low to high resolutions. ESOD~\cite{liu2024esod} uses patch slicing and sparse prediction to reduce redundant computation on high-resolution features. DM-EFS~\cite{sharma2025dm} expands feature sets to enrich small-object representations. These works show the value of high-resolution cues for tiny objects. However, many detection pipelines still discard the highest-resolution backbone feature to keep computation affordable, leaving the finest local details underused. In contrast, our method revives this discarded feature through semantic enhancement, making its fine-grained details more compatible and usable for tiny object detection.

\subsection{Auxiliary supervision and label assignment}
Auxiliary supervision has been widely used to strengthen intermediate representations in object detection. In CNN-based detectors, DSFPN~\cite{yang2019dually}, LAF-YOLOv10~\cite{farooqui2026laf} and Jin et al.~\citep{jin2022you} attach auxiliary prediction heads to early backbone stages or bottom-up pathways to improve feature learning. A similar idea is adopted in DETR-like models~\cite{zhang2022dino, huang2024dq, dome-detr, peng2024dfine, huang2024deim}, where auxiliary losses on intermediate decoder layers are commonly used to stabilize training and facilitate optimization. Beyond intermediate decoder supervision, MLP-DINO~\cite{cao2024mlp} introduces query-independent category supervision to mitigate the sensitivity of category predictions to bounding box prediction. 

Label assignment is another key factor affecting detection performance. ATSS~\cite{zhang2020bridging} shows that positive and negative sample definitions are critical. For generic object detection, AutoAssign~\cite{zhu2020autoassign} and IQDet~\cite{ma2021iqdet} select high-quality regions according to predicted quality scores, while FCOS~\cite{tian2019fcos} formulates detection as dense per-pixel prediction. For transformer-based detectors, DS-Det~\cite{cao2025dsdet} introduces a unified Single-Query paradigm and disentangles cross-attention-based box localization with one-to-many assignment from self-attention-based deduplication with one-to-one assignment. However, tiny objects make assignment more difficult because small localization shifts can strongly affect matching quality. Recent methods therefore redesign assignment for tiny objects. UGS~\cite{sun2025uncertainty} shows that IoU-based assignment can be unstable for tiny objects and proposes a classification-based objective. DCNet~\cite{zhang2025tiny} uses dynamic scale-aware assignment, while RFLA~\cite{rfla} matches Gaussian receptive fields with ground truth to alleviate positive-sample scarcity. Unlike these methods that mainly design assignment for final prediction, our work focuses on strengthening intermediate representations. Specifically, we propose a geometry-adaptive assignment strategy and auxiliary supervision to build coarse classification and localization priors for tiny objects, while the auxiliary head is removed during inference.

\section{Method}
\label{method}
\subsection{Overall framework}


\begin{figure}[t] 
  \centering
  \includegraphics[width=1.0\linewidth]{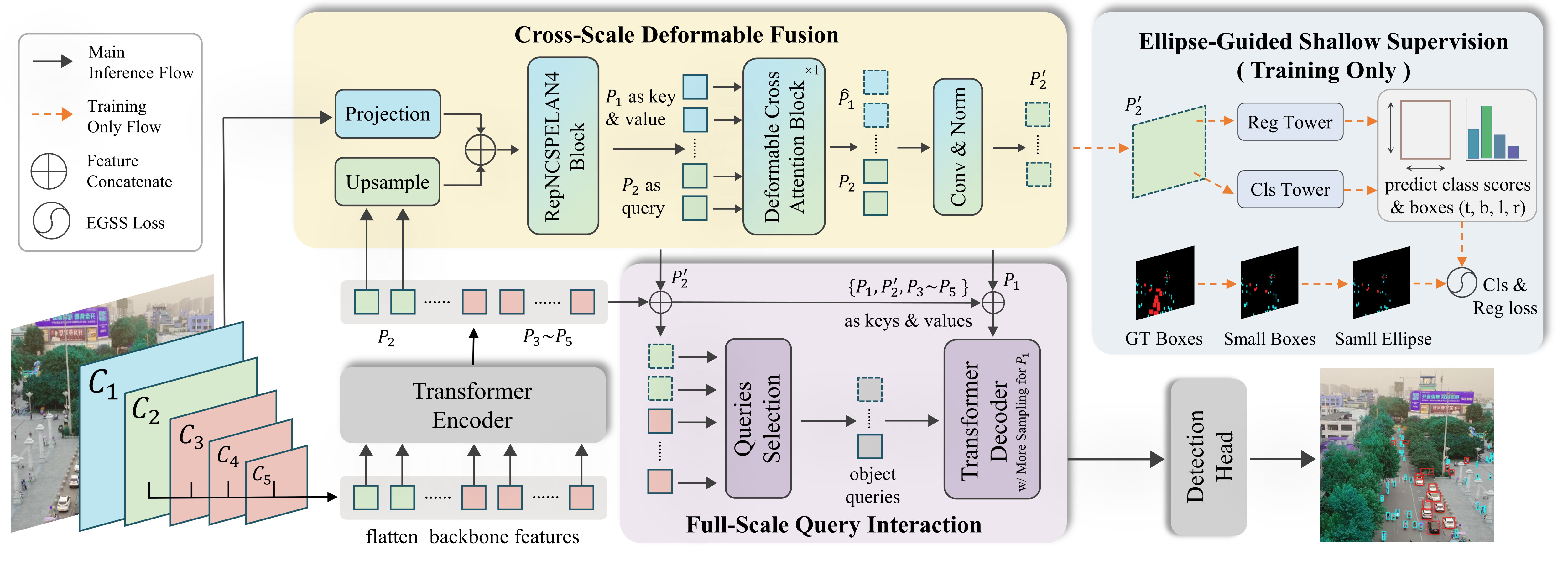}

    \caption{
    Overview of ReDHF. CSDF revives the discarded highest-resolution feature $C_1$ as a semantically enhanced feature $P_1$ and injects its fine-grained details into $P_2$ to obtain the detail-enhanced shallow feature $P_2'$. EGSS further supervises $P_2'$ with ellipse-guided auxiliary supervision during training, while FSQI exposes $P_1$, $P_2'$, and deeper features to decoder queries for final detection.
    }

  \label{fig:overall_architecture}
\end{figure}

Given an input image, let $\{C_i\}_{i=1}^{l}$ denote the multi-level backbone features, and let $\{P_i\}_{i=1}^{l}$ denote the corresponding multi-scale features after encoder context modeling, where $C_1$ is the highest-resolution feature. To revive the discarded highest-resolution feature without incurring excessive computation or amplifying background responses, we propose \textbf{ReDHF}, a plug-and-play framework for transformer-based tiny object detection. As shown in Fig.~\ref{fig:overall_architecture}, ReDHF consists of three components: Cross-Scale Deformable Fusion (CSDF, Sec.~\ref{subsec:csdf}), Ellipse-Guided Shallow Supervision (EGSS, Sec.~\ref{subsec:egss}), and Full-Scale Query Interaction (FSQI, Sec.~\ref{subsec:fsqi}). During training, CSDF and FSQI are optimized end-to-end in the main detection pipeline, while EGSS supervises the shallow feature to learn stable classification and localization priors. During inference, EGSS is completely removed, and only CSDF and FSQI remain for feature construction and decoder interaction.

\subsection{Cross-scale deformable fusion}
\label{subsec:csdf}

CSDF is designed to exploit the fine-grained details of the highest-resolution backbone feature while reducing the side effect of its noisy background responses. Instead of directly using raw $C_1$, CSDF first constructs a semantically enhanced highest-resolution feature $P_1$, and then injects its high-resolution details into the shallow feature $P_2$.

\paragraph{Semantically enhanced highest-resolution feature construction.}
Directly using $C_1$ is suboptimal because it contains rich details but weak semantics~\cite{FPN}, making it sensitive to background noises. To address this issue, we inject the semantic information from $P_2$ into $C_1$ through a lightweight RepNCSPELAN4 block~\cite{peng2024dfine}. Since lower layers preserve local details while higher layers aggregate more global context~\cite{raghu2021vision}, we only upsample $P_2$ and fuse it with the projected $C_1$ as follows:
\begin{equation}
P_1 = \mathcal{F}_{\text{RepNCSPELAN4}}
\big(
\mathrm{Concat}(\mathrm{Up}(P_2), \phi(C_1))
\big),
\end{equation}
where $\phi(\cdot)$ denotes channel projection and $\mathcal{F}_{\text{RepNCSPELAN4}}$ denotes the fusion block. In this way, $P_1$ preserves the high-resolution local details of $C_1$ while inheriting the semantic prior from $P_2$, alleviating the semantic gaps between different depth features.

\paragraph{High-resolution detail injection for tiny objects.}
After obtaining $P_1$, we further use it to compensate for the fine-grained details that are missing in the shallow feature $P_2$. However, either a direct dense convolution or a bidirectional cross attention for the fusion of $P_1$ and $P_2$ will cause computational bottlenecks due to the large spatial size of $P_1$. We therefore adopt a deformable cross-attention design, where $P_2$ serves as query and $P_1$ serves as key and value:
\begin{equation}
\hat{P}_1 = \mathrm{DeformAttn}(Q=P_2, K=P_1, V=P_1).
\end{equation}
Here, $\hat{P}_1$ denotes the retrieved detail feature, which is adaptively extracted from $P_1$ according to the spatial prior of $P_2$. Since $\hat{P}_1$ is essentially local detail responses from $P_1$, it cannot replace the original $P_2$. Instead, we fuse it back into $P_2$ to form the final detail-enhanced shallow feature $P_2'$:
\begin{equation}
P_2' =
\mathcal{F}_{\text{act}}
\Big(
\mathcal{F}_{\text{norm}}
\big(
\mathrm{Conv}_{1 \times 1}
(\mathrm{Concat}(P_2,\hat{P}_1))
\big)
\Big),
\end{equation}
where $\mathcal{F}_{\text{norm}}$ and $\mathcal{F}_{\text{act}}$ denote normalization and activation, respectively. The resulting $P_2^ {\prime}$ retains the semantic content of $P_2$ and the local detail supplement from the highest-resolution $P_1$. Compared with the global or deformable self-attention interaction, the computational cost of CSDF is more acceptable, as analyzed in Appendix~\ref{app:efficiency_analysis}, making the use of the highest-resolution feature more practical in tiny object detection.

\subsection{Ellipse-guided shallow supervision}
\label{subsec:egss}

EGSS aims to guide $P_2'$ to learn reliable foreground classification and coarse localization priors, rather than making final predictions on the intermediate representation. Although CSDF enriches $P_2'$ with high-resolution details, it may also transfer dense background responses from $P_1$ to $P_2$. As shown in Fig.~\ref{fig:featuresMaps}, both $P_1$ and $P_2$ can respond strongly to background textures and local patterns. When injecting detail cues from $P_1$ into $P_2$, these background responses may be transferred together with object-related details, thereby amplifying non-object responses in complex scenes. To address this side effect, we introduce EGSS, which applies shallow auxiliary supervision to $P_2'$ during training and is completely removed during inference, introducing no additional inference cost.

\paragraph{Ellipse-guided label assignment.}
Standard FCOS-style dense supervision~\cite{tian2019fcos} usually adopts box-shaped assignment or fixed-radius center sampling, which is not ideal for tiny objects. Since tiny objects occupy only a few pixels, box boundaries often include background pixels and introduce noisy supervision. Center sampling can reduce boundary noise, but a fixed region cannot adapt well to different aspect ratios and may even remove all valid positives for extremely tiny objects.

At the same time, the detail-enhanced shallow feature $P_2'$ has a limited receptive field and is more suitable for detecting small objects. For large objects, a feature point on $P_2'$ often perceives only a small part of the instance and cannot understand the complete structure well. Assigning such objects to supervise $P_2'$ may introduce ambiguous signals and distract the shallow feature from tiny-object regions. Motivated by the scale-aware design of FCOS~\cite{tian2019fcos}, we restrict EGSS to size-matched objects and define positive samples with a scale filtering strategy and an ellipse-guided geometric constraint. Let $\mathcal{G}$ denote the set of ground-truth boxes, $\mathcal{P}$ denote the set of feature locations on $P_2'$, and $s_2$ denote the stride of $P_2'$. The positive sample set $\mathcal{A}_{\text{pos}}$ is defined as:
\begin{equation}
\begin{aligned}
\mathcal{A}_{pos} = \Big\{ (p_i, q_j) \mid \;
p_i = (x_i, y_i) \in \mathcal{P}, \
q_j = (c_x^{(j)}, c_y^{(j)}, w_j, h_j) \in \mathcal{G}, \\
r_{\min} \le \frac{\sqrt{w_j h_j}}{s_2} \le r_{\max}, \
\frac{(x_i - c_x^{(j)})^2}{(w_j/2)^2}
+ \frac{(y_i - c_y^{(j)})^2}{(h_j/2)^2}
\le 1
\Big\}.
\end{aligned}
\end{equation}
Here, $(c_x^{(j)}, c_y^{(j)}, w_j, h_j)$ denote the center, width, and height of the $j$-th ground-truth box, and $[r_{\min}, r_{\max}]$ filters objects whose sizes do not match the receptive field of $P_2'$. Compared with rectangular assignment, the elliptical constraint naturally adapts better to object aspect ratios and suppresses boundary noise for tiny objects, while the scale filter focuses supervision on size-matched tiny objects, which helps $P_2'$ pay more attention to learning to perceive tiny objects. Meanwhile, when a feature point matches multiple objects, we assign it to the smallest one to reduce ambiguity and bias the supervision toward smaller instances.

\paragraph{Shallow auxiliary supervision.}

Based on the above assignment, we attach a lightweight FCOS-style~\cite{tian2019fcos} auxiliary head to $P_2'$ during training. The head contains two parallel towers for classification and regression, respectively. Each tower consists of two $3 \times 3$ convolution layers, followed by a prediction layer that outputs class logits or box regression results. The auxiliary loss is:
\begin{equation}
\mathcal{L}_{aux}
=
\lambda_{cls}\mathcal{L}_{cls}
+
\lambda_{reg}\mathcal{L}_{reg},
\end{equation}
where $\lambda_{cls}$ and $\lambda_{reg}$ are the loss weights. We use focal loss for $\mathcal{L}_{cls}$ to handle the severe 
foreground-background imbalance and a combination of $\ell_1$ loss and GIoU~\cite{rezatofighi2019giou} loss for $\mathcal{L}_{reg}$ to provide more stable regression supervision.

Tiny-object regression is naturally sensitive to small localization shifts~\cite{sun2025uncertainty}, especially on noisy shallow features. Since $P_2'$ still contains non-negligible background responses, as shown in Fig.~\ref{fig:featuresMaps}, a strong regression loss may force the auxiliary branch to fit unstable box targets and disturb foreground discrimination. We therefore assign a small regression weight. In addition, we do not adopt centerness supervision by default, since tiny objects have only a few valid locations on the feature map and their centerness targets can vary sharply and become unreliable on the noisy shallow feature. EGSS therefore focuses on foreground classification and coarse localization, while final box refinement is left to the decoder. We empirically validate this design choice in Appendix~\ref{app:loss_weight}.

\subsection{Full-scale query interaction}
\label{subsec:fsqi}

In DETR-like detectors~\cite{carion2020end, zhu2020deformable, liu2022dabdetr}, final predictions are generated by decoder queries. Instead of attending only to high-level features, FSQI enables these queries to directly access the fine-grained details in $P_1$, further improving tiny object detection performance. Here, ``full-scale'' means that the decoder can attend to all feature levels, from the highest-resolution $P_1$ to deeper semantic features.

\paragraph{Full-scale encoder memory.}
We flatten and concatenate the semantically enhanced highest-resolution feature $P_1$, the detail-enhanced shallow feature $P_2'$, and the higher-level features $\{P_3,\dots,P_l\}$ to construct the full-scale encoder memory:
\begin{equation}
\mathcal{M}
=
\mathrm{Concat}
\big(
\mathrm{Flatten}(P_1),
\mathrm{Flatten}(P_2'),
\dots,
\mathrm{Flatten}(P_l)
\big).
\end{equation}
Here, $\mathcal{M}$ serves as the value tokens for decoder cross-attention. Since $P_1$ is semantically enhanced from raw $C_1$, its tokens are more compatible with the encoded features $\{P_3,\dots,P_l\}$, alleviating the distribution gap between different-level features. To avoid noisy shallow responses disturbing query selection, FSQI excludes $P_1$ from query initialization and exposes it only during decoder cross-attention, where decoder queries can adaptively retrieve relevant high-resolution details.

\paragraph{Scale-aware sparse sampling.}
Since the receptive field of $P_1$ is relatively small, if we keep its sampling setting the same as other feature levels, decoder queries may fail to collect enough information from $P_1$. We therefore allocate more sampling points on $P_1$ in decoder deformable cross-attention phase. In practice, FSQI increases the number of sampling points on $P_1$ from 4 to 8, while keeping it unchanged for the remaining feature levels. This gives queries a larger local search radius on $P_1$ and improves their access to fine-grained spatial cues for tiny object detection.

\section{Experiments}

\subsection{Datasets and evaluation metrics}

\paragraph{Datasets.}
We evaluate ReDHF on three challenging tiny-object detection datasets: AI-TOD-v2~\cite{Aitod-v2}, VisDrone~\cite{zhu2021visdrone}, and SODA-D~\cite{SODA}. AI-TOD-v2 and VisDrone cover dense aerial scenes with many tiny instances, while SODA-D provides high-resolution traffic scenes with many small and extremely small objects. These datasets cover both aerial and driving scenarios, making them suitable for evaluating the generality of ReDHF. Detailed dataset statistics are provided in Appendix~\ref{app:dataset_details}.

\paragraph{Evaluation metrics.}

We use Average Precision (AP)~\cite{lin2014microsoft} as the main metric, which is calculated as the mean AP over the range of IoU thresholds from 0.50 to 0.95, with an interval of 0.05. To better evaluate performance on tiny objects, we also report dataset-specific scale-aware metrics. For AI-TOD-v2~\cite{Aitod-v2}, we follow~\cite{AI-TOD_2020_ICPR} and report AP$_{vt}$, AP$_t$, AP$_s$, and AP$_m$. For VisDrone~\cite{zhu2021visdrone} and SODA-D~\cite{SODA}, we report AP$_s$ and AP$_{eS}$ as the tiny-object metrics, respectively. Full scale-aware metrics are provided in Appendix~\ref{app:full_vis_sodad_results}.


\begin{table}[t]
  \caption{Experiments on AI-TOD-v2 test split. All models are evaluated with 800×800 input resolution. We compare ReDHF with both CNN-based and Transformer-based detectors. $\dagger$ denotes our re-implementation of the corresponding baseline. $\star$ denotes the average GFLOPs measured over the whole split when the computational cost varies across images.}
  \label{tab:main_aitod}
    \centering
    \renewcommand{\arraystretch}{1.1} 
    \setlength{\tabcolsep}{1.5mm} 
    \footnotesize
    \begin{tabular}{l c c c c c c c c c}
        \toprule[1.2pt] 
        \textbf{Method} & \textbf{\#Params.} & \textbf{GFLOPs} & \textbf{AP} & \textbf{AP$_{50}$} & \textbf{AP$_{75}$}  & \textbf{AP$_{vt}$} & \textbf{AP$_{t}$} & \textbf{AP$_{s}$}& \textbf{AP$_{m}$} \\
        \midrule
        
        \rowcolor[HTML]{F3EAE1} 
        \multicolumn{10}{l}{\textit{CNN-based models}} \\

        Cascade R-CNN \cite{cai18cascadercnn} & 68.9M & 88.7 & 15.1 & 34.2 & 11.2 & 0.1 & 11.5 & 26.7 & 38.5 \\
        DetectoRS \cite{detectors} & 134.8M & $\approx$ 160 & 16.1 & 35.5 & 12.5 & 0.1 & 12.6 & 28.3 & 40.0 \\
        DetectoRS w/ RFLA  \cite{rfla} & -- & -- & 24.8 & 55.2 & 18.5 & 9.3 & 24.8 & 30.3 & 38.2 \\
        YOLOv12-X  \cite{tian2025yolov12} & 59.3M & 185.4 & 16.1 & 33.5 & 13.4 & 5.4 & 17.0 & 21.4 & 28.5 \\
        
        \midrule
        
        \rowcolor[HTML]{F3EAE1} 
        \multicolumn{10}{l}{\textit{Transformer-based models}} \\
        DNTR \cite{dntr} & 128.4M & 298.1 & 26.2 & 56.7 & 20.2 & 12.8 & 26.4 & 31.0 & 37.0 \\
        Deformable-DETR\cite{zhu2020deformable} & 40M & 232.5 & 18.2 & 59.7 & 10.2 & 5.8 & 16.8 & 25.2 & 34.0 \\
        
        DINO-4scale$\dagger$  \cite{zhang2022dino} & 47M & 354.4 & 25.1 & 60.8 & 16.3 & 12.1 & 24.0 & 31.0 & 40.5 \\
        DQ-DETR$\dagger$ \cite{huang2024dq} & 58.7M & 1805.4$^\star$ & 30.2 & 68.6 & 22.3 & 15.3 & 30.5 & 36.5 & 44.6 \\
        
        Dome-DETR-S$\dagger$ \cite{dome-detr} & 13.2M & 173.1$^\star$ & 32.1  & 66.3 & 26.9 & 17.5 & 32.1 & 38.2 & 45.0 \\
        Dome-DETR-M$\dagger$ \cite{dome-detr} & 23.9M & 252.6$^\star$ & 34.0  & 68.8 & 29.4 & 18.9 & 34.1 & 38.9 & 47.4 \\
        Dome-DETR-L$\dagger$ \cite{dome-detr} & 36.0M & 358.7$^\star$ & 34.6  & 69.2 & 30.4 & 19.6 & 35.6 & \textbf{39.8} & 47.3 \\
        
        \midrule
        \rowcolor[HTML]{F3EAE1} 
        \multicolumn{10}{l}{\textit{Our method}} \\
        \rowcolor{black!3}
        DINO-4scale \textcolor{red}{w/ ReDHF} & 49.7M  & 407.8 & 26.0\textbf{\textcolor[HTML]{FF0000}{\tiny +0.9}} & 62.1 & 17.3 & 13.8\textbf{\textcolor[HTML]{FF0000}{\tiny +1.7}} & 25.2 & 32.0 & 39.6\\
        \rowcolor{black!6}
        DQ-DETR \textcolor{red}{w/ ReDHF} & 61.9M  & 1872.3$^\star$ & 31.4\textbf{\textcolor[HTML]{FF0000}{\tiny +1.2}} & 69.8 & 22.6 & 16.6\textbf{\textcolor[HTML]{FF0000}{\tiny +1.3}} & 31.2 & 36.7 & 44.3\\
        \rowcolor{black!9}
        Dome-DETR-S \textcolor{red}{w/ ReDHF} & 15.9M  & 188.2$^\star$ & 33.5\textbf{\textcolor[HTML]{FF0000}{\tiny +1.4}} & 68.1 & 28.9 & 18.9\textbf{\textcolor[HTML]{FF0000}{\tiny +1.4}} & 33.5 & 38.7 & 45.8\\
        \rowcolor{black!12}
        Dome-DETR-M \textcolor{red}{w/ ReDHF} & 26.7M  & 304.7$^\star$ & 35.0\textbf{\textcolor[HTML]{FF0000}{\tiny +1.0}} & 68.9 & 30.6 & 20.7\textbf{\textcolor[HTML]{FF0000}{\tiny +1.8}} & 34.8 & 39.2 & \textbf{47.7}\\
        \rowcolor{black!15}
        Dome-DETR-L \textcolor{red}{w/ ReDHF} & 39.0M  & 415.3$^\star$ & \textbf{35.7\textcolor[HTML]{FF0000}{\tiny +1.1}} & \textbf{70.2} & \textbf{31.5} & \textbf{20.8\textcolor[HTML]{FF0000}{\tiny +1.2}} & \textbf{35.9} & \textbf{39.8} & 47.4\\
        
        \bottomrule[1.2pt] 
    \end{tabular}
\end{table}

\begin{table}[t]
\centering
\caption{
Experiments on VisDrone validation split and SODA-D test split. Each entry reports baseline $\rightarrow$ ReDHF, with the gain over the corresponding baseline shown in parentheses. AP$_s$ is used as the tiny-object metric for VisDrone, and AP$_{eS}$ is used for SODA-D. 
}
\label{tab:compact_vis_sodad}
\setlength{\tabcolsep}{4.2pt}
\renewcommand{\arraystretch}{1.08}
\footnotesize
\resizebox{0.95\linewidth}{!}{%
\begin{tabular}{l c c c c}
\toprule
\textbf{Baseline}
& \textbf{VisDrone AP}
& \textbf{VisDrone AP$_s$}
& \textbf{SODA-D AP}
& \textbf{SODA-D AP$_{eS}$} \\
\midrule

DINO-4scale~\cite{zhang2022dino}
& 34.3 $\rightarrow$ 35.4 {\scriptsize \textcolor[HTML]{FF0000}{(+1.1)}}
& 25.6 $\rightarrow$ 27.4 {\scriptsize \textcolor[HTML]{FF0000}{(+1.8)}}
& 28.6 $\rightarrow$ 30.1 {\scriptsize \textcolor[HTML]{FF0000}{(+1.5)}}
& 10.5 $\rightarrow$ 12.6 {\scriptsize \textcolor[HTML]{FF0000}{(+2.1)}} \\

DQ-DETR~\cite{huang2024dq}
& 35.7 $\rightarrow$ 37.1 {\scriptsize \textcolor[HTML]{FF0000}{(+1.4)}}
& 28.2 $\rightarrow$ 29.6 {\scriptsize \textcolor[HTML]{FF0000}{(+1.4)}}
& 34.4 $\rightarrow$ 35.5 {\scriptsize \textcolor[HTML]{FF0000}{(+1.1)}}
& 17.0 $\rightarrow$ 18.5 {\scriptsize \textcolor[HTML]{FF0000}{(+1.5)}} \\

Dome-DETR-S~\cite{dome-detr}
& 35.9 $\rightarrow$ 37.5 {\scriptsize \textcolor[HTML]{FF0000}{(+1.6)}}
& 28.3 $\rightarrow$ 30.3 {\scriptsize \textcolor[HTML]{FF0000}{(+2.0)}}
& 35.4 $\rightarrow$ 36.8 {\scriptsize \textcolor[HTML]{FF0000}{(+1.4)}}
& 17.6 $\rightarrow$ 19.6 {\scriptsize \textcolor[HTML]{FF0000}{(+2.0)}} \\

Dome-DETR-M~\cite{dome-detr}
& 38.9 $\rightarrow$ 40.2 {\scriptsize \textcolor[HTML]{FF0000}{(+1.3)}}
& 31.4 $\rightarrow$ \textbf{33.1} {\scriptsize \textcolor[HTML]{FF0000}{(+1.7)}}
& 37.2 $\rightarrow$ 38.4 {\scriptsize \textcolor[HTML]{FF0000}{(+1.2)}}
& 18.6 $\rightarrow$ 20.4 {\scriptsize \textcolor[HTML]{FF0000}{(+1.8)}} \\

Dome-DETR-L~\cite{dome-detr}
& 39.0 $\rightarrow$ \textbf{40.6} {\scriptsize \textcolor[HTML]{FF0000}{(+1.6)}}
& 31.5 $\rightarrow$ \textbf{33.1} {\scriptsize \textcolor[HTML]{FF0000}{(+1.6)}}
& 37.6 $\rightarrow$ \textbf{38.7} {\scriptsize \textcolor[HTML]{FF0000}{(+1.1)}}
& 19.7 $\rightarrow$ \textbf{21.4} {\scriptsize \textcolor[HTML]{FF0000}{(+1.7)}} \\

\midrule
\textit{Avg. gain}
& \textbf{\textcolor[HTML]{FF0000}{+1.4}}
& \textbf{\textcolor[HTML]{FF0000}{+1.7}}
& \textbf{\textcolor[HTML]{FF0000}{+1.3}}
& \textbf{\textcolor[HTML]{FF0000}{+1.8}} \\

\bottomrule
\end{tabular}%
}
\end{table}

\subsection{Implementation details}
\label{imple_detail}
We evaluate ReDHF on three representative baselines: DINO-4scale~\cite{zhang2022dino}, DQ-DETR~\cite{huang2024dq}, and Dome-DETR~\cite{dome-detr}. Dome-DETR uses HGNetv2~\cite{zhao2024detrs} as the backbone, while DINO-4scale and DQ-DETR use ResNet-50~\cite{he2016deep}. All experiments are trained end-to-end and evaluated on 4 Tesla V100 GPUs. Unless otherwise stated, we keep all the baseline architectures unchanged and only introduce the three components of our method. For Dome-DETR and DQ-DETR, we follow their original configurations on AI-TOD-v2 and VisDrone. Since SODA-D is not included in their original benchmarks, we reuse the VisDrone training configuration for this dataset. For DINO-4scale, we adopt its default COCO~\cite{lin2014microsoft} 24-epoch training recipe for all three datasets, changing only dataset-specific settings. For EGSS, we set $r_{\min}=0$ for all datasets, with $r_{\max}=8$ for AI-TOD-v2 and $r_{\max}=16$ for VisDrone and SODA-D. Complete training and implementation details are provided in Appendix~\ref{app:implementation}.




\subsection{Main results}
\label{main_results}
\paragraph{AI-TOD-v2.}

Table~\ref{tab:main_aitod} reports the results on the AI-TOD-v2~\cite{Aitod-v2} test split. ReDHF consistently improves all evaluated DETR-like variants, with average gains of +1.1 AP and +1.5 AP$_{vt}$. Dome-DETR-L w/ ReDHF achieves the best result of 35.7 AP, surpassing previous state-of-the-art methods. The improvement is more notable on very tiny objects, with an average gain of +1.5 AP$_{vt}$ across the three baseline variants. ReDHF brings consistent gains for most metrics except for AP$_s$ or AP$_m$, which is expected since medium objects rely less on fine-grained local details and can already be well represented by deeper semantic features.

\paragraph{VisDrone and SODA-D.}

Table~\ref{tab:compact_vis_sodad} reports results on VisDrone~\cite{zhu2021visdrone} and SODA-D~\cite{SODA}. ReDHF improves all baselines on both datasets, with average gains of +1.4 AP / +1.7 AP$_s$ on VisDrone and +1.3 AP / +1.8 AP$_{eS}$ on SODA-D. The best variants achieve 40.6 AP on VisDrone and 38.7 AP on SODA-D, respectively. Full scale-aware results are provided in Appendix~\ref{app:full_vis_sodad_results}.

\subsection{Efficiency analysis}
\label{subsec:efficiency}


ReDHF aims to improve tiny object detection with moderate overhead. As shown in Table~\ref{tab:main_aitod}, it consistently improves DETR-like baselines with limited increases in model size and computation. For example, on Dome-DETR-S~\cite{dome-detr}, ReDHF improves AP by +1.4 while just increasing the average GFLOPs from 173.1 to 188.2. Similar trends are observed on DINO-4scale~\cite{zhang2022dino}, DQ-DETR~\cite{huang2024dq}, and other Dome-DETR variants. Appendix~\ref{app:efficiency_analysis} provides a detailed complexity analysis.

\subsection{Ablation study}

We conduct ablation studies on AI-TOD-v2~\cite{Aitod-v2} to analyze the effectiveness of ReDHF and its key design choices. Unless otherwise stated, Dome-DETR-S~\cite{dome-detr} is used as the default baseline. Additional ablations on sampling points, and EGSS loss design are provided in Appendix~\ref{app:additional_ablations}.

\paragraph{Main ablation experiment.}

Table~\ref{tab:ablation_study} reports the contribution of each component. For CSDF, raw $C_1$ reduces AP from 32.1 to 31.1, while the semantically aligned $P_1$ improves AP to 32.7, indicating that raw high-resolution details suffer from background responses and semantic mismatch. For FSQI, adding raw $C_1$ to the encoder memory improves AP to 32.5 since decoder queries can selectively attend to useful high-resolution details and are therefore less sensitive to shallow noise, and replacing it with $P_1$ further improves AP to 32.8. In contrast, using $P_1$ for query initialization drops AP to 31.9, suggesting that shallow high-resolution features are unsuitable for initializing decoder queries. When cooperating CSDF and FSQI together, AP can increase to 33.1, and EGSS further improves the full model to 33.5 AP and 18.9 AP$_{vt}$, confirming the complementary effects of all components.

\begin{table}[h]
\centering

\begin{minipage}[t]{0.59\linewidth}
\vspace{0pt}
\centering

\caption{
Component ablation of ReDHF. $C_1$ and $P_1$ denote the raw highest-resolution feature and its semantically enhanced one. ``q. init.'' means query initialization with $P_1$.
}

\label{tab:ablation_study}
\setlength{\tabcolsep}{2.3pt}
\renewcommand{\arraystretch}{1.05}
\scalefont{0.6}
\resizebox{\linewidth}{!}{
\begin{tabular}{cc ccc c ccc}
\toprule
\multicolumn{2}{c}{CSDF} & \multicolumn{3}{c}{FSQI} & \multirow{2}{*}{EGSS} 
& \multirow{2}{*}{AP} & \multirow{2}{*}{AP$_{vt}$} & \multirow{2}{*}{AP$_t$} \\
\cmidrule(r){1-2} \cmidrule(lr){3-5}
$C_1$ & $P_1$ & $C_1$ & $P_1$ & q. init. & & & & \\
\midrule
 & & & & & & 32.1 & 17.5 & 32.1 \\
\midrule
\checkmark & & & & & & 31.1 & 16.8 & 30.8 \\
& \checkmark & & & & & 32.7 & 18.0 & 32.7 \\
\midrule
& & \checkmark & & & & 32.5 & 17.8 & 32.4 \\
& & & \checkmark & \checkmark & & 31.9 & 17.6 & 32.0 \\
& & & \checkmark & & & 32.8 & 18.1 & 32.8 \\

\midrule
& \checkmark & & \checkmark & & & 33.1 & 18.3 & 33.0 \\
& \checkmark & & \checkmark & & \checkmark & \textbf{33.5} & \textbf{18.9} & \textbf{33.5} \\
\bottomrule
\end{tabular}
}
\end{minipage}
\hfill
\begin{minipage}[t]{0.38\linewidth}
\vspace{0pt}
\centering

\caption{Comparison of different label assignment strategies for EGSS. The best result is highlighted in bold.}
\label{tab:egss_assignment}
\setlength{\tabcolsep}{4pt}
\renewcommand{\arraystretch}{1.05}
\scriptsize
\resizebox{\linewidth}{!}{
\begin{tabular}{lccc}
\toprule
Assignment & AP & AP$_{50}$ & AP$_{vt}$ \\
\midrule
Rectangle & 33.2 & 66.9 & 17.6 \\
Circle & 33.3 & 67.8 & 18.1 \\
Ellipse (ours) & \textbf{33.5} & \textbf{68.1} & \textbf{18.9}  \\
\bottomrule
\end{tabular}
}

\vspace{5.6mm}

\caption{Ablation on the detail injection level.}
\label{tab:inject_level}
\setlength{\tabcolsep}{4pt}
\renewcommand{\arraystretch}{1.05}
\scriptsize
\resizebox{\linewidth}{!}{
\begin{tabular}{lccc}
\toprule
Inject layer & AP & AP$_{50}$ & AP$_{vt}$ \\
\midrule
$P_2 \sim P_3$ & 32.8 & 66.5 & 18.0 \\
$P_2 \sim P_5$ & 32.4 & 66.3 & 18.3 \\
$P_2$ only & \textbf{33.5} & \textbf{68.1} & \textbf{18.9} \\
\bottomrule
\end{tabular}
}

\end{minipage}

\end{table}

\paragraph{Design ablations.}
Tables~\ref{tab:egss_assignment} and~\ref{tab:inject_level} study two key design choices of ReDHF. For EGSS, Rectangle, Circle, and Ellipse denote box-based, short-axis circle-based, and ellipse-guided assignment, respectively. Our method achieves the best performance, improving AP$_{vt}$ by +1.3 over Rectangle and +0.8 over Circle. For CSDF, injecting high-resolution details only into $P_2$ gives the best result, reaching 33.5 AP and 18.9 AP$_{vt}$. Full results and analysis are provided in Appendix~\ref{app:label_assign} and Appendix~\ref{app:detail_inject}.

\subsection{Visualization analysis}
\begin{figure}[t]
  \centering
  \includegraphics[width=1.0\linewidth]{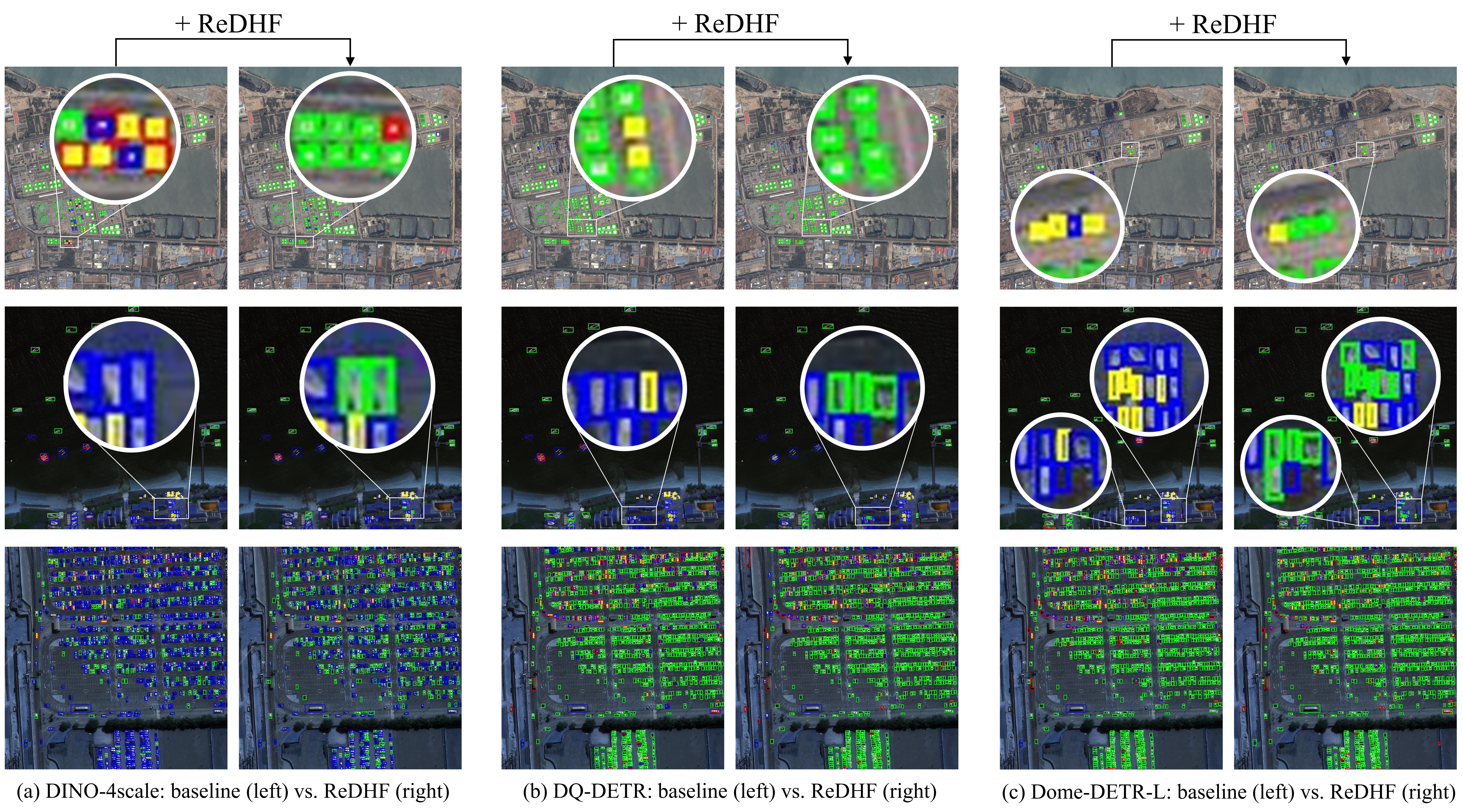}
    \caption{
Qualitative comparison on the AI-TOD-v2 test split. Each pair compares a baseline detector with its ReDHF variant on the same scene. The \textcolor{green}{green}, \textcolor{red}{red}, \textcolor{yellow}{yellow}, and \textcolor{blue}{blue} boxes represent TP, FP, FN with ground-truth box short side $\leq 4$ pixels, and the remaining FN, respectively. Compared with the baselines, ReDHF produces cleaner predictions and helps recover more tiny objects, with more green boxes and fewer red/yellow/blue boxes. Class labels are omitted for clarity.
}
    
  \label{fig:prediction}
\end{figure}

\paragraph{Detection visualization.}
Fig.~\ref{fig:prediction} compares ReDHF with corresponding baselines across different scenarios on AI-TOD-v2~\cite{Aitod-v2}. ReDHF makes better predictions and recovers more ultra-tiny objects, showing that our method helps detectors better preserve and use fine-grained details for tiny objects.


\paragraph{Feature maps visualization.}
\begin{figure}[htbp]
  \centering
  \includegraphics[width=1.0\linewidth]{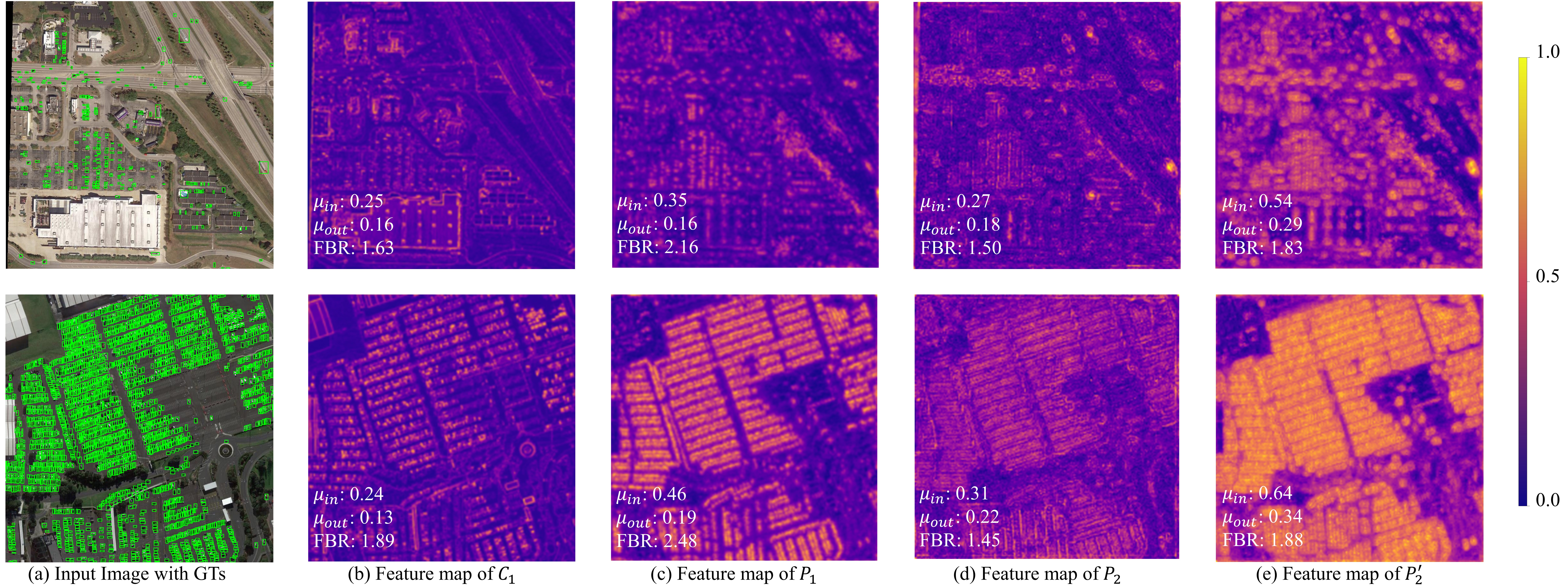}
  
\caption{
Feature visualization and response analysis of CSDF. We compare $C_1$ vs. $P_1$ and $P_2$ vs. $P_2'$. The bottom-left statistics report $\mu_{\text{in}}$, $\mu_{\text{out}}$, and FBR defined in Eq.~\ref{eq:fbr_stats}. After CSDF enhancement, $P_1$ and $P_2'$ show stronger response concentration on GT regions.
}
  \label{fig:featuresMaps}
\end{figure}

To further analyze CSDF, we compute channel-wise $\ell_2$ response maps and measure their mean responses inside and outside GT regions, denoted as $\mu_{\text{in}}$ and $\mu_{\text{out}}$. The foreground-to-background response ratio (FBR) is then defined as:
\begin{equation}
\label{eq:fbr_stats}
\mu_{\text{in}}=
\frac{1}{|\Omega_{\text{in}}|}\sum_{(x,y)\in\Omega_{\text{in}}}R(x,y), \quad 
\mu_{\text{out}}=
\frac{1}{|\Omega_{\text{out}}|}\sum_{(x,y)\in\Omega_{\text{out}}}R(x,y), 
\quad
\mathrm{FBR}=\frac{\mu_{\mathrm{in}}}{\mu_{\mathrm{out}}}.
\end{equation}
Here, $R(x,y)$ denotes the normalized response at location $(x,y)$, while $\Omega_{\mathrm{in}}$ and $\Omega_{\mathrm{out}}$ denote the set of locations inside and outside GT regions. A larger FBR indicates stronger response concentration on object regions. As shown in Fig.~\ref{fig:featuresMaps}, $P_1$ achieves higher FBR than raw $C_1$, and $P_2'$ shows the same improvement over $P_2$. This suggests that CSDF makes the highest-resolution feature more object-focused and transfers more effective fine-grained details to the shallow feature.

\section{Conclusion}
\label{sec:conclusion}

In this paper, we propose ReDHF, a plug-and-play framework that revives the discarded highest-resolution feature for tiny object detection. Motivated by the contradiction between the limited representation of tiny objects and the high cost of utilizing high-resolution features, we introduce CSDF, EGSS, and FSQI to effectively address these issues, achieving consistent improvements across different DETR-like models on the AI-TOD-v2, VisDrone and SODA-D datasets. 
Although retaining the highest-resolution representation introduces additional computation, ReDHF yields consistent gains that are most pronounced on tiny objects, and we only observe localized degradation for large objects in dense scenes with smaller-capacity variants. Detailed analyses are provided in Appendix~\ref{app:limitations}. Future work will explore more adaptive high-resolution feature utilization under low-resource and crowded scenes. We hope ReDHF can inspire continued progress in this area.


\begin{ack}



This work was supported by the Pengcheng Laboratory Research Project (PCL2025A14), the Guangdong Basic and Applied Basic Research Foundation (Grant No. 2024A1515011292, Project No. 2024A1515011437, Project No. 2025B1515120012), and the National Natural Science Foundation of China (Grant No. 62606254). The authors declare no competing interests.




\end{ack}



\bibliography{references}      





\newpage
\section*{Appendix}

\appendix


This appendix provides additional implementation details, complete experimental results, and supplementary analyses for ReDHF.
Section~\ref{app:implementation} describes the training and method-specific configurations.
Section~\ref{app:dataset_details} provides detailed statistics of the three benchmarks, while Section~\ref{app:full_vis_sodad_results} reports the complete results on VisDrone and SODA-D.
Section~\ref{app:efficiency_analysis} analyzes the practical efficiency and cost-accuracy trade-off of high-resolution feature reuse.
Section~\ref{app:broader_analysis} examines performance across object scales and generalization to a general-purpose detection benchmark.
Section~\ref{app:limitations} discusses the practical limitations and failure cases of ReDHF.
Finally, Section~\ref{app:additional_ablations} provides additional ablations and robustness analyses of the key design choices.

\section{Additional Implementation Details}
\label{app:implementation}

\paragraph{Baselines and backbones.}
We evaluate ReDHF on three representative DETR-based detectors: DINO-4scale~\citep{zhang2022dino}, DQ-DETR~\citep{huang2024dq}, and Dome-DETR~\citep{dome-detr}. Dome-DETR adopts HGNetv2~\citep{zhao2024detrs} as its backbone, while DINO-4scale and DQ-DETR use ResNet-50~\citep{he2016deep}. All experiments are trained end-to-end and evaluated on 4 Tesla V100 GPUs. Unless otherwise stated, we keep all the baseline architectures unchanged and only introduce the three components of our method. 

\paragraph{Training configurations.}
For Dome-DETR and DQ-DETR, we follow their original training configurations on AI-TOD-v2 and VisDrone. Since SODA-D is not included in their original benchmarks, we use the corresponding VisDrone training configuration for SODA-D. For DINO-4scale, we follow its default COCO~\citep{lin2014microsoft} 24-epoch training recipe on all three datasets, while adapting only the dataset-specific settings.

\paragraph{ReDHF settings.}
For EGSS, we set $r_{\min}=0$ on all datasets. We use $r_{\max}=8$ on AI-TOD-v2 and $r_{\max}=16$ on VisDrone and SODA-D. The default auxiliary-loss weights are $\lambda_{\mathrm{cls}}=1.0$ and $\lambda_{\mathrm{reg}}=0.10$, and centerness supervision is not used. For the deformable interactions, we use 8 sampling points by default in both CSDF and FSQI. The sensitivity analyses of the sampling-point number and the EGSS loss configuration are provided in Sec.~\ref{app:sampling_points} and Sec.~\ref{app:loss_weight}, respectively.

\section{Dataset details}
\label{app:dataset_details}

\paragraph{AI-TOD-v2.}
AI-TOD-v2~\cite{Aitod-v2} contains 28,036 aerial images with 752,745 annotated object instances. The dataset is split into 11,214 training images, 2,804 validation images, and 14,018 test images. It is characterized by extremely small object sizes, with an average object size of only 12.7 pixels. In particular, 86\% of the objects are smaller than 16 pixels, and the largest object does not exceed 64 pixels. The object density also varies greatly across images, ranging from 1 to 2,667 objects per image, with an average of 24.64 and a standard deviation of 63.94.

\paragraph{VisDrone.}
VisDrone~\cite{zhu2021visdrone} contains 14,018 images across 10 distinct categories, with 6,471 images for training, 548 for validation, and 3,190 for testing. It covers diverse objects such as pedestrians, bicycles, and vehicles, and presents severe challenges in terms of diverse crowd densities. Specifically, the average number of objects per image is 40.7, with a standard deviation of 46.41.

\paragraph{SODA-D.}
SODA-D~\cite{SODA} contains 24,828 high-quality traffic-scene images with 278,433 annotated instances across 9 categories. The training, validation, and test subsets are split with an approximate ratio of 50\% : 20\% : 30\%. The dataset includes high-resolution driving scenes with many small and extremely small objects under diverse real-world conditions, including different viewpoints, weather, and illumination. These properties make SODA-D a challenging benchmark for small object detection in driving scenarios.

\section{Full results on VisDrone and SODA-D}
\label{app:full_vis_sodad_results}


\paragraph{VisDrone.}
Table~\ref{tab:main_visdrone} reports the results on the VisDrone~\cite{zhu2021visdrone} validation split. Our ReDHF on DQ-DETR~\cite{huang2024dq} improves AP by  +1.4\%. On Dome-DETR-S and Dome-DETR-M, it raises AP from 35.9 to 37.5 and from 38.9 to 40.2, respectively. When applied to Dome-DETR-L, ReDHF achieves the best AP(40.6\%), improving AP, AP$_{50}$, and AP$_{75}$ by +1.6\%, +1.7\%, and +2.2\%, respectively.

\begin{table}[t]
  \caption{Experiments on VisDrone val split. We compare ReDHF with both CNN-based and Transformer-based detectors. $\dagger$ denotes our re-implementation of the corresponding baseline. $\star$ denotes the average GFLOPs measured over the whole split when the computational cost varies across images.}
  \label{tab:main_visdrone}
    \centering
    \renewcommand{\arraystretch}{1.1} 
    \setlength{\tabcolsep}{1.2mm} 
    \scalefont{0.85}   
    \begin{tabular}{l c c c c c c c c c}
        \toprule[1.2pt] 
        \textbf{Method} & \textbf{\#Params.} & \textbf{GFLOPs} & \textbf{Image Scale} & \textbf{AP} & \textbf{AP$_{50}$} & \textbf{AP$_{75}$}  & \textbf{AP$_{s}$} & \textbf{AP$_{m}$} & \textbf{AP$_{l}$} \\
        \midrule
        
        \rowcolor[HTML]{F3EAE1} 
        \multicolumn{10}{l}{\textit{CNN-based models}} \\

        YOLOv11 ~\cite{khanam2024yolov11} & 25.3M & 86.6 & 640×640 & 27.9 & -- & -- & 18.1 & 40.9 & 53.1 \\
        YOLOv12 ~\cite{tian2025yolov12} & 26.3M & 88.6 & 640×640 & 29.2 & -- & -- & 18.7 & 41.8 & 51.7 \\
        RFLA ~\cite{rfla} & 69.2M & 162.0 & 800×800 & 24.3 & -- & -- & 14.4 & 35.9 & 45.8 \\
        YOLC ~\cite{liu2024yolc} & -- & -- & 1024×640 & 39.6 & -- & -- & 32.9 & 49.3 & 56.2 \\
        CFINet ~\cite{CFINet} & 41.4M & 291.9 & 1200×1200 & 23.9 & -- & -- & 16.3 & 34.7 & 57.4 \\
        ESOD ~\cite{liu2024esod} & -- & 232.9 & 1920×1920 & 39.4 & 62.9 & -- & 31.1 & -- & -- \\

        \midrule
        
        \rowcolor[HTML]{F3EAE1} 
        \multicolumn{10}{l}{\textit{Transformer-based models}} \\
        DDQ DETR ~\cite{DDQ} & -- & 270 & 1333×800 & 31.6 & -- & -- & 23.2 & 42.1 & 51.1 \\
        DPAKS ~\cite{xu2026dpaks} & 218.3M & 1457.6 & 1333×800 & 39.1 & -- & -- & 32.1 & 49.5 & \textbf{63.4} \\
        UGS ~\cite{sun2025uncertainty} & -- & -- & 1333×800 & 38.1 & 61.9 & -- & 24.2 & 47.3 & 56.6 \\
        DINO-4scale$\dagger$  \cite{zhang2022dino} & 47M & 544.3 & 1333×800 & 34.3 & 57.3 & 34.8 & 25.6 & 45.6 & 59.5 \\
        DQ-DETR$\dagger$ \cite{huang2024dq} & 58.7M & 2826.6$^\star$ & 1333×800  & 35.7 & 59.4 & 36.4 & 28.2 & 45.8 & 55.1 \\
        Dome-DETR-S$\dagger$ \cite{dome-detr} & 13.2M & 167.3$^\star$ & 800×800 & 35.9 & 58.0 & 37.2 & 28.3 & 45.9 & 58.9 \\
        Dome-DETR-M$\dagger$ \cite{dome-detr} & 23.9M & 296.2$^\star$ & 800×800 & 38.9 & 62.1 & 40.7 & 31.4 & 49.3 & 62.3 \\
        Dome-DETR-L$\dagger$ \cite{dome-detr} & 36.0M & 385.1$^\star$ & 800×800 & 39.0 & 62.3 & 40.6 & 31.5 & 49.3 & 62.8 \\
        
        \midrule
        \rowcolor[HTML]{F3EAE1} 
        \multicolumn{10}{l}{\textit{Our method}} \\

        \rowcolor{black!3}
        DINO-4scale \textcolor{red}{w/ ReDHF} & 49.7M  & 627.6 & 1333×800 & 35.4\textbf{\textcolor[HTML]{FF0000}{\tiny +1.1}} & 58.1 & 36.2 & 27.4\textbf{\textcolor[HTML]{FF0000}{\tiny +1.8}} & 45.6 & 59.4\\
        \rowcolor{black!6}
        DQ-DETR \textcolor{red}{w/ ReDHF} & 61.9M & 2928.3$^\star$ & 1333×800 & 37.1\textbf{\textcolor[HTML]{FF0000}{\tiny +1.4}}  & 61.0 & 37.9 & 29.6\textbf{\textcolor[HTML]{FF0000}{\tiny +1.4}} & 46.5 & 57.1\\
        \rowcolor{black!9}
        Dome-DETR-S \textcolor{red}{w/ ReDHF} & 15.9M & 180.8$^\star$ & 800×800 & 37.5\textbf{\textcolor[HTML]{FF0000}{\tiny +1.6}} & 59.4 & 39.5 & 30.3\textbf{\textcolor[HTML]{FF0000}{\tiny +2.0}} & 47.1 & 58.6\\
        \rowcolor{black!12}
        Dome-DETR-M \textcolor{red}{w/ ReDHF} & 26.7M & 338.3$^\star$ & 800×800 & 40.2\textbf{\textcolor[HTML]{FF0000}{\tiny +1.3}} & 63.3 & 42.6 & 33.1\textbf{\textcolor[HTML]{FF0000}{\tiny +1.7}} & 50.0 & 62.0\\
        \rowcolor{black!15}
        Dome-DETR-L \textcolor{red}{w/ ReDHF} & 39.0M & 449.3$^\star$ & 800×800 & \textbf{40.6\textcolor[HTML]{FF0000}{\tiny +1.6}} & \textbf{64.0} & \textbf{42.8} & \textbf{33.1\textcolor[HTML]{FF0000}{\tiny +1.6}} & \textbf{51.0} & 63.1\\
        
        \bottomrule[1.2pt] 
    \end{tabular}
\end{table}


\paragraph{SODA-D.}
Table~\ref{tab:main_sodad} reports the results on the SODA-D~\cite{SODA} test split. ReDHF brings +1.5\% AP on DINO-4scale, +1.1\% AP on DQ-DETR, and +1.4\%, +1.2\%, +1.1\% AP on Dome-DETR-S/M/L, respectively. The gains are more pronounced on extremely small objects. Dome-DETR-L w/ ReDHF achieves the best result, with 38.7 AP and 21.4 AP$_{eS}$.

\begin{table}
  \caption{Experiments on SODA-D test split. We compare ReDHF with both CNN-based and Transformer-based detectors. $\dagger$ denotes our re-implementation of the corresponding baseline. $\star$ denotes the average GFLOPs measured over the whole split when the computational cost varies across images.}
  \label{tab:main_sodad}
    \centering
    \renewcommand{\arraystretch}{1.1} 
    \setlength{\tabcolsep}{1.2mm} 
    \scalefont{0.75}   
    \begin{tabular}{l c c c c c c c c c c}
        \toprule[1.2pt] 
        \textbf{Method} & \textbf{\#Params.} & \textbf{GFLOPs} & \textbf{Image Scale} & \textbf{AP} & \textbf{AP$_{50}$} & \textbf{AP$_{75}$}  & \textbf{AP$_{eS}$} & \textbf{AP$_{rS}$} & \textbf{AP$_{gs}$} & \textbf{AP$_{N}$} \\
        \midrule
        
        \rowcolor[HTML]{F3EAE1} 
        \multicolumn{11}{l}{\textit{CNN-based models}} \\

        Faster RCNN~\cite{ren2015faster} & 41.2M & 292.3 & 1200×1200 & 28.9 & 59.7 & 24.2 & 13.9 & 25.6 & 34.3 & 43.2 \\
        Cascade RCNN~\cite{cai18cascadercnn} & 68.9M & 320.1 & 1200×1200 & 31.2 & 59.9 & 27.8 & 14.1 & 27.5 & 37.1 & 46.9 \\
        RFLA~\cite{rfla} & 41.2M & 292.1 & 1200×1200 & 29.7 & 60.2 & 25.2 & 13.2 & 26.9 & 35.4 & 44.6 \\

        \midrule
        
        \rowcolor[HTML]{F3EAE1} 
        \multicolumn{11}{l}{\textit{Transformer-based models}} \\
        Deformable-DETR\cite{zhu2020deformable} & 35.2M & 739.1 & 1200×1200 & 19.2 & 44.8 & 13.7 & 6.3 & 15.4 & 24.9 & 34.2 \\

        DINO-4scale$\dagger$  \cite{zhang2022dino} & 47M & 354.4 & 800×800 & 28.6 & 59.3 & 24.2 & 10.5 & 24.2 & 36.2 & 47.4 \\
        DQ-DETR$\dagger$ \cite{huang2024dq} & 58.7M & $1802.6^\star$ & 800×800  & 34.4 & 62.6 & 31.8 & 17.0 & 30.9 & 41.2 & 51.4 \\
        Dome-DETR-S$\dagger$ \cite{dome-detr} & 13.2M & 160.12 & 800×800 & 35.4 & 63.9 & 33.0 & 17.6 & 31.8 & 42.1 & 52.2 \\
        Dome-DETR-M$\dagger$ \cite{dome-detr} & 23.9M & 291.78 & 800×800 & 37.2 & 66.6 & 34.7 & 18.6 & 34.1 & 44.8 & 53.6 \\
        Dome-DETR-L$\dagger$ \cite{dome-detr} & 36.0M & 373.12 & 800×800 & 37.6 & 67.5 & 35.2 & 19.7 & 34.1 & 44.5 & 54.8 \\
        
        \midrule
        \rowcolor[HTML]{F3EAE1} 
        \multicolumn{11}{l}{\textit{Our method}} \\
        \rowcolor{black!3}
        DINO-4scale \textcolor{red}{w/ ReDHF} & 49.7M  & 407.8 & 800×800 & 30.1\textbf{\textcolor[HTML]{FF0000}{\tiny +1.5}} & 59.6 & 26.3 & 12.6\textbf{\textcolor[HTML]{FF0000}{\tiny +2.1}} & 25.6 & 37.3 & 47.6 \\
        \rowcolor{black!6}
        DQ-DETR \textcolor{red}{w/ ReDHF} & 61.9M & 1867.5 & 800×800 & 35.5\textbf{\textcolor[HTML]{FF0000}{\tiny +1.1}} & 63.6 & 33.3 & 18.5\textbf{\textcolor[HTML]{FF0000}{\tiny +1.5}} & 32.2 & 41.9 & 51.4 \\
        \rowcolor{black!9}
        Dome-DETR-S \textcolor{red}{w/ ReDHF} & 15.9M & 184.08 & 800×800 & 36.8\textbf{\textcolor[HTML]{FF0000}{\tiny +1.4}} & 65.7 & 34.5 & 19.6\textbf{\textcolor[HTML]{FF0000}{\tiny +2.0}} & 33.6 & 43.2 & 52.8 \\
        \rowcolor{black!12}
        Dome-DETR-M \textcolor{red}{w/ ReDHF} & 26.7M & 332.34 & 800×800 & 38.4\textbf{\textcolor[HTML]{FF0000}{\tiny +1.2}} & 67.9 & 35.8 & 20.4\textbf{\textcolor[HTML]{FF0000}{\tiny +1.8}} & 34.9 & 45.1 & 54.6 \\
        \rowcolor{black!15}
        Dome-DETR-L \textcolor{red}{w/ ReDHF} & 39.0M & 435.78 & 800×800 & 38.7\textbf{\textcolor[HTML]{FF0000}{\tiny +1.1}} & 68.4 & 36.3 & 21.4\textbf{\textcolor[HTML]{FF0000}{\tiny +1.7}} & 35.1 & 45.4 & 55.0 \\
        
        \bottomrule[1.2pt] 
    \end{tabular}
\end{table}

\begin{table}[h]
\centering
\caption{
Computational cost comparison on DINO-4scale. We report GFLOPs of the backbone, encoder, decoder, other parts, and the whole model. ``Direct $C_1$'' denotes directly adding the raw highest-resolution feature into the encoder and decoder. ``Others'' includes components not counted in the backbone, encoder, or decoder.
}
\label{tab:app_efficiency_dino}
\footnotesize
\setlength{\tabcolsep}{5pt}
\renewcommand{\arraystretch}{1.10}
\begin{tabular}{lccccc}
\toprule
Model & Backbone & Encoder & Decoder & Others & Whole \\
\midrule
DINO-4scale~\cite{zhang2022dino} & 104.5 & 203.9 & 34.9 & 11.1 & 354.4 \\
DINO-4scale w/ direct $C_1$ & 104.5{\color{red}\small(--)} & 2709.5{\color{red}\small(+2505.6)} & 161.0{\color{red}\small(+126.1)} & 82.7{\color{red}\small(+71.6)} & 3057.7{\color{red}\small(+2703.3)} \\
DINO-4scale w/ ReDHF & 104.5{\color{red}\small(--)} & 251.2{\color{red}\small(+47.3)} & 40.7{\color{red}\small(+5.8)} & 11.4{\color{red}\small(+0.3)} & 407.8{\color{red}\small(+53.4)} \\
\bottomrule

\end{tabular}
\end{table}

\section{Detailed efficiency analysis}
\label{app:efficiency_analysis}

\subsection{Computational cost of high-resolution feature reuse}

Let $N_i$ denote the number of tokens in $P_i$, and $K$ denote the number of sampling points. Standard global self attention for $P_i$ and $P_{i+1}$ has a complexity of $\mathcal{O}(N_i^2 \cdot N_{i+1}^2)$, while that in deformable self attention is $\mathcal{O}(K \cdot N_i + K \cdot N_{i+1})$, which becomes expensive when the spatial resolution is large. In contrast, the complexity of our deformable interactions in CSDF is $\mathcal{O}(K \cdot N_{i+1})$, where $K \ll N_i$, making the use of the highest-resolution feature $P_1$ much more practical. Table~\ref{tab:app_efficiency_dino} compares the computational cost of different ways to use the highest-resolution feature on DINO-4scale~\citep{zhang2022dino}. Directly adding raw $C_1$ into the encoder and decoder causes a large cost increase, raising the whole-model GFLOPs from 354.4 to 3057.7. The main cost comes from the encoder, whose GFLOPs increase from 203.9 to 2709.5 due to dense interaction on the highest-resolution feature. The decoder cost also increases from 34.9 to 161.0. This is because each decoder layer performs value projection on the multi-scale memory, and the very large spatial size of $C_1$ makes repeated projection expensive.

In contrast, ReDHF increases the whole-model GFLOPs only to 407.8. Compared with direct $C_1$ usage, ReDHF reduces the total computation by about 86.7\%, the encoder computation by about 90.7\%, and the decoder computation by about 74.7\%. 
This reduction comes from two designs. First, CSDF retrieves high-resolution details through sparse deformable interaction instead of dense global interaction. Second, FSQI exposes $P_1$ to decoder cross-attention in a lightweight manner. In our implementation, the value projection for the highest-resolution feature is applied only once, while the remaining feature levels follow the original per-layer projection. Although FSQI allocates more sampling points to $P_1$, the total GFLOPs increase remains moderate, as shown in Fig.~\ref{fig:ablation_points_gflops}. Moreover, EGSS is only used during training and removed during inference, so it does not add test-time computation. Overall, ReDHF is an efficient high-resolution detail reintroduction framework for tiny object detection. It avoids the expensive computational cost of utilizing raw $C_1$, while still effectively releasing the potential of the highest-resolution feature for tiny object detection.

\subsection{Compute-matched comparison with input resolution scaling}

One misconception for the improvement of ReDHF is that it introduces additional computation for high-resolution feature processing. To distinguish the contribution of the proposed design from simply increasing the computational budget, we compare ReDHF with resolution-scaled baselines while keeping the remaining detector and training settings unchanged.

\begin{table}[t]
\caption{
Compute-matched comparison with input-resolution scaling on AI-TOD-v2.
We increase only the input resolution of the baselines while keeping the detector
and training settings unchanged.
The best result within each baseline family is highlighted in bold.
}
\label{tab:resolution_control}

\setlength{\tabcolsep}{5pt}
\renewcommand{\arraystretch}{1.10}
\centering
\begin{tabular}{lcccc}
\toprule
Model & Input & GFLOPs & AP & AP$_{vt}$ \\
\midrule
DINO-4scale & 800 & 354.4 & 25.1 & 12.1 \\
DINO-4scale & 864 & 412.4 & 25.4 & 12.4 \\
DINO-4scale w/ ReDHF & 800 & 407.8 & \textbf{26.0} & \textbf{13.8} \\
\midrule
Dome-DETR-L & 800 & 358.7 & 34.6 & 19.6 \\
Dome-DETR-L & 880 & 481.1 & 35.0 & 20.0 \\
Dome-DETR-L w/ ReDHF & 800 & 415.3 & \textbf{35.7} & \textbf{20.8} \\
\bottomrule

\end{tabular}
\end{table}

As shown in Table~\ref{tab:resolution_control}, increasing the input resolution of DINO-4scale from 800 to 864 raises the computation from 354.4 to 412.4 GFLOPs, but improves AP and AP$_{vt}$ by only 0.3 points. In contrast, ReDHF achieves 26.0 AP and 13.8 AP$_{vt}$ at 407.8 GFLOPs, outperforming the resolution-scaled baseline by 0.6 AP and 1.4 AP$_{vt}$ under a comparable computational budget. A similar trend is observed for Dome-DETR-L. These results indicate that the gains of ReDHF cannot be reproduced by uniformly increasing the input resolution, supporting the effectiveness of explicitly recovering and exploiting the discarded high-resolution representation.

\subsection{Inference latency and GPU memory}

GFLOPs do not fully characterize the practical cost of retaining the highest-resolution feature. We therefore measure wall-clock inference latency and GPU memory under identical baseline and ReDHF settings. All measurements are conducted on a Tesla V100 with batch size 1.

\begin{table}[t]
\caption{
Practical inference cost of ReDHF on AI-TOD-v2 and VisDrone.
Wall-clock latency and GPU memory are measured on a single Tesla V100 with batch size 1
under identical baseline and ReDHF settings.
Each entry reports baseline $\rightarrow$ ReDHF.
Tiny AP denotes AP$_{vt}$ on AI-TOD-v2 and AP$_s$ on VisDrone.
}
\label{tab:latency_memory}
\setlength{\tabcolsep}{7pt}
\renewcommand{\arraystretch}{1.12}
\centering
{\fontsize{7.7pt}{8.7pt}\selectfont
\begin{tabular}{llcccc}
\toprule
Dataset & Variant &
Tiny AP $\uparrow$ &
Latency (ms) $\downarrow$ &
Avg. memory (GiB) $\downarrow$ &
Peak memory (GiB) $\downarrow$ \\
\midrule
\multirow{3}{*}{AI-TOD-v2}
& Dome-DETR-S & 17.5 $\rightarrow$ \textbf{18.9} & 409.6 $\rightarrow$ 507.6 & 0.30 $\rightarrow$ 0.56 & 0.63 $\rightarrow$ 1.72 \\
& Dome-DETR-M & 18.9 $\rightarrow$ \textbf{20.7} & 450.0 $\rightarrow$ 578.0 & 0.39 $\rightarrow$ 0.65 & 0.77 $\rightarrow$ 1.86 \\
& Dome-DETR-L & 19.6 $\rightarrow$ \textbf{20.8} & 487.8 $\rightarrow$ 655.2 & 0.47 $\rightarrow$ 0.76 & 0.86 $\rightarrow$ 1.99 \\
\midrule
\multirow{3}{*}{VisDrone}
& Dome-DETR-S & 28.3 $\rightarrow$ \textbf{30.3} & 454.5 $\rightarrow$ 584.1 & 0.27 $\rightarrow$ 0.46 & 0.63 $\rightarrow$ 1.72 \\
& Dome-DETR-M & 31.4 $\rightarrow$ \textbf{33.1} & 512.8 $\rightarrow$ 680.1 & 0.36 $\rightarrow$ 0.57 & 0.77 $\rightarrow$ 1.86 \\
& Dome-DETR-L & 31.5 $\rightarrow$ \textbf{33.1} & 543.5 $\rightarrow$ 751.8 & 0.47 $\rightarrow$ 0.69 & 0.88 $\rightarrow$ 2.01 \\
\bottomrule
\end{tabular}
}
\end{table}

Table~\ref{tab:latency_memory} shows that ReDHF introduces additional latency and memory consumption due to retaining and querying $P_1$. Across the evaluated Dome-DETR variants on AI-TOD-v2 and VisDrone, the additional latency ranges from 98.0 to 208.3 ms, while the increase in average GPU memory ranges from 0.19 to 0.29 GiB. Meanwhile, ReDHF consistently improves the corresponding tiny-object AP by 1.2-2.0 points. These results quantify the practical cost associated with recovering high-resolution evidence and complement the GFLOPs analysis above. Overall, ReDHF achieves a favorable accuracy-efficiency trade-off, where the additional runtime and memory are accompanied by consistent improvements in tiny-object detection.


\section{Detection Quality and Generalization}
\label{app:broader_analysis}

\subsection{Scale-wise analysis}

Because ReDHF is designed primarily for tiny objects, we further examine whether recovering high-resolution evidence introduces a systematic trade-off at larger object scales. Table~\ref{tab:scale_analysis} summarizes the average gains over the five evaluated DETR variants, with the scale categories ordered from the smallest to larger objects for each dataset.

The improvements are consistently largest on the smallest object categories and gradually decrease with object size. Nevertheless, the averaged gains on the remaining scales are non-negative across all three datasets. This indicates that ReDHF concentrates its benefits on objects most affected by feature downsampling without systematically sacrificing the detection quality of larger objects.

\begin{table}
\caption{
Scale-wise performance gains of ReDHF averaged over the five evaluated DETR variants.
The dataset-specific metrics are ordered from smaller to larger object scales.
}
\label{tab:scale_analysis}
\setlength{\tabcolsep}{5pt}
\renewcommand{\arraystretch}{1.12}
\centering
\begin{tabular}{lccc}
\toprule
Dataset & Smallest objects & Intermediate objects & Larger objects \\
\midrule
AI-TOD-v2
& AP$_{vt}$ +1.48
& AP$_t$ +0.86
& AP$_s$ +0.40,\; AP$_m$ +0.00 \\
VisDrone
& AP$_s$ +1.70
& AP$_m$ +0.86
& AP$_l$ +0.32 \\
SODA-D
& AP$_{eS}$+1.82
& AP$_{rS}$ +1.26
& AP$_{gS}$ +0.82,\; AP$_N$ +0.40 \\
\bottomrule
\end{tabular}
\end{table}

\subsection{Generalization on COCO}

To further examine whether ReDHF remains effective beyond tiny-object-focused benchmarks, we evaluate DINO-4scale with ReDHF on COCO using the same training recipe as the baseline. As shown in Table~\ref{tab:coco_generalization}, ReDHF improves the overall AP from 49.0 to 49.6. Improvements are also observed across all object scales, with AP$_s$, AP$_m$, and AP$_l$ increasing by 0.6, 0.8, and 0.5 points, respectively. These results suggest that recovering high-resolution evidence does not require sacrificing medium- or large-object performance and that ReDHF can also provide complementary information on a general-purpose detection benchmark.

\begin{table}
\caption{
Generalization of ReDHF on COCO using DINO-4scale.
Both models use the same 12-epoch training recipe.
}
\label{tab:coco_generalization}
\setlength{\tabcolsep}{7pt}
\renewcommand{\arraystretch}{1.12}
\centering
\begin{tabular}{lcccc}
\toprule
Model & AP & AP$_s$ & AP$_m$ & AP$_l$ \\
\midrule
DINO-4scale & 49.0 & 32.0 & 52.3 & 63.0 \\
DINO-4scale w/ ReDHF &
\textbf{49.6 (+0.6)} &
\textbf{32.6 (+0.6)} &
\textbf{53.1 (+0.8)} &
\textbf{63.5 (+0.5)} \\
\bottomrule
\end{tabular}
\end{table}

\section{Limitations Analysis}
\label{app:limitations}

Retaining and querying an additional highest-resolution feature $P_1$ inevitably introduces extra computation and memory compared with the original detectors. As quantified in Table~\ref{tab:app_efficiency_dino}, however, this overhead is substantially smaller than that of directly incorporating the raw highest-resolution feature. More importantly, the additional cost is accompanied by consistent improvements in tiny-object detection, indicating a favorable accuracy-efficiency trade-off for exploiting high-resolution evidence.

As analysed in Sec.~\ref{app:broader_analysis}, the benefits of ReDHF are concentrated on tiny objects, and the average improvement gradually decreases as object size increases, although no systematic degradation is observed at larger scales. This behavior is consistent with the motivation of ReDHF: high-resolution evidence is most useful for objects whose visual information is severely affected by feature downsampling. To identify where the occasional degradation on larger objects occurs, we divide the VisDrone validation set into two equally sized subsets according to the number of annotated instances per image and evaluate the same trained checkpoints on both subsets.

\begin{table}[h]
\caption{
Density-wise analysis on the VisDrone validation set.
Images are ranked by the number of annotated instances and divided into two equally
sized sparse and dense subsets.
Each entry reports baseline $\rightarrow$ ReDHF, with the absolute change in parentheses.
}
\label{tab:density_analysis}
\setlength{\tabcolsep}{2.0pt}
\renewcommand{\arraystretch}{1.12}
\centering
{\fontsize{5.6pt}{8.7pt}\selectfont
\begin{tabular}{lcccccc}
\toprule
Variant &
Sparse AP$_s$ &
Sparse AP$_m$ &
Sparse AP$_l$ &
Dense AP$_s$ &
Dense AP$_m$ &
Dense AP$_l$ \\
\midrule
Dome-DETR-S &
28.24 $\rightarrow$ 30.31 (\textbf{+2.07}) &
47.09 $\rightarrow$ 48.91 (\textbf{+1.82}) &
60.60 $\rightarrow$ 63.12 (\textbf{+2.52}) &
28.11 $\rightarrow$ 30.58 (\textbf{+2.47}) &
44.76 $\rightarrow$ 45.96 (\textbf{+1.20}) &
57.15 $\rightarrow$ 55.60 (-1.55) \\
Dome-DETR-M &
30.86 $\rightarrow$ 33.02 (\textbf{+2.16}) &
50.80 $\rightarrow$ 51.10 (\textbf{+0.30}) &
64.86 $\rightarrow$ 65.13 (\textbf{+0.27}) &
32.02 $\rightarrow$ 33.46 (\textbf{+1.44}) &
48.46 $\rightarrow$ 49.41 (\textbf{+0.95}) &
60.76 $\rightarrow$ 59.80 (-0.96) \\
Dome-DETR-L &
31.54 $\rightarrow$ 32.90 (\textbf{+1.36}) &
50.12 $\rightarrow$ 52.09 (\textbf{+1.97}) &
68.34 $\rightarrow$ 68.86 (\textbf{+0.52}) &
31.71 $\rightarrow$ 33.37 (\textbf{+1.66}) &
49.15 $\rightarrow$ 50.40 (\textbf{+1.25}) &
55.60 $\rightarrow$ 56.62 (\textbf{+1.02}) \\
\bottomrule
\end{tabular}
}
\end{table}

As shown in Table~\ref{tab:density_analysis}, across all three variants, ReDHF improves all object scales in sparse scenes. In dense scenes, AP$_s$ and AP$_m$ remain consistently improved, whereas AP$_l$ decreases for Dome-DETR-S and Dome-DETR-M but increases for Dome-DETR-L. Dense scenes contain many spatially adjacent high-resolution responses, which may increase background interference and hinder smaller-capacity detectors from preserving the broader semantic context required for large-object recognition. Therefore, the degradation is specific to large objects in dense scenes for smaller-capacity detectors, rather than a systematic trade-off across object scales.

\section{Additional ablations and Robustness analyses}
\label{app:additional_ablations}

\subsection{Label assignment strategy in EGSS}
\label{app:label_assign}
Table~\ref{tab:app_egss_assignment} compares different label assignment strategies for EGSS. The results show that the positive sample definition is important for shallow auxiliary supervision. Rectangle-based assignment reaches 33.2 AP, but its AP$_{vt}$ is 17.6. This suggests that using the whole box region as positives may introduce boundary background pixels, which is harmful for very tiny objects.

We further compare a short-axis circle-based strategy, where the positive region is a circle centered at the object center and its radius is set to the short semi-axis of the corresponding ground-truth box. This strategy reduces boundary noise and improves AP$_{vt}$ from 17.6 to 18.1, but it still ignores object aspect ratios and may discard valid positive regions along the long axis.

In contrast, ellipse-guided assignment adapts to the object geometry by using both the long and short axes. It achieves the best AP, AP$_{50}$, AP$_{75}$, AP$_{vt}$, AP$_t$, and AP$_s$. Compared with rectangle-based and short-axis circle-based assignment, it improves AP$_{vt}$ by +1.3 and +0.8, respectively. These results show that the elliptical constraint better suppresses boundary noise while preserving sufficient valid positives for tiny objects. Although the circle-based strategy obtains the best AP$_m$, EGSS mainly targets tiny and small objects, where ellipse-guided assignment gives more consistent gains. Therefore, we adopt ellipse-guided assignment as the default strategy.

\begin{table}[t]
\caption{Comparison of different label assignment strategies for EGSS. The best result in each column is highlighted in bold.}
\label{tab:app_egss_assignment}
\centering

\setlength{\tabcolsep}{6pt}
\renewcommand{\arraystretch}{1.12}
\begin{tabular}{l c c c c c c c}
\toprule
Assignment strategy & AP & AP$_{50}$ & AP$_{75}$ & AP$_{vt}$ & AP$_t$ & AP$_s$ & AP$_m$ \\
\midrule
Rectangle-based & 33.2 & 66.9 & 28.7 & 17.6 & 33.0 & 38.2 & 45.0 \\
Short-axis circle-based & 33.3 & 67.8 & 28.4 & 18.1 & 32.9 & 38.4 & \textbf{46.1} \\
Ellipse-guided (ours) & \textbf{33.5} & \textbf{68.1} & \textbf{28.9} & \textbf{18.9} & \textbf{33.5} & \textbf{38.7} & 45.8 \\
\bottomrule
\end{tabular}
\end{table}

\subsection{Detail injection level analysis}
\label{app:detail_inject}
Table~\ref{tab:app_inject_level} studies where the highest-resolution details from $P_1$ should be injected. All settings keep the other components unchanged and only vary which feature levels receive the highest-resolution details. Injecting details only into $P_2$ gives the best result, reaching 33.5 AP and 18.9 AP$_{vt}$. This suggests that $P_2$ is the most suitable level to receive highest-resolution details, because it is spatially closest to $P_1$ while still containing more semantic information than raw shallow features. In contrast, injecting details into more levels does not bring further gains. When the injection range is extended to $P_2\sim P_3$ or $P_2\sim P_5$, AP decreases to 32.8 and 32.4, respectively. This is likely because deeper feature levels are more responsible for medium and large objects, while excessive high-resolution details may become redundant or even introduce noise into semantic features. These results support our design choice of injecting high-resolution details only into $P_2$.

\begin{table}[t]
\caption{
Ablation on the detail injection level. We keep other components unchanged and only vary the target feature levels. The best result in each column is highlighted in bold.
}
\label{tab:app_inject_level}

\centering
\setlength{\tabcolsep}{6pt}
\renewcommand{\arraystretch}{1.15}
\begin{tabular}{l c c c c c c c}
\toprule
Inject layer & AP & AP$_{50}$ & AP$_{75}$ & AP$_{vt}$ & AP$_t$ & AP$_s$ & AP$_m$ \\
\midrule
$P_2 \sim P_3$ & 32.8 & 66.5 & 28.1 & 18.0 & 32.7 & 38.5 & 45.0 \\
$P_2 \sim P_5$ & 32.4 & 66.3 & 27.3 & 18.3 & 32.2 & 38.2 & 44.6 \\
$P_2$ only     & \textbf{33.5} & \textbf{68.1} & \textbf{28.9} & \textbf{18.9} & \textbf{33.5} & \textbf{38.7} & \textbf{45.8} \\
\bottomrule
\end{tabular}
\end{table}

\subsection{Downsampling strategy for semantically enhanced highest-resolution feature}
Table~\ref{tab:p1_downsampling} compares different strategies for adapting $P_1$ to the resolution of $P_2$ in CSDF. The results show that simple fixed-rule downsampling is not always effective. Bilinear interpolation severely degrades performance, suggesting that directly resizing high-resolution responses may blur tiny-object details and introduce background interference. AvgPool and strided convolution perform better, but they still remain inferior to deformable cross-attention on the main metrics.

Deformable cross-attention achieves the best AP and AP$_{vt}$. This is because it allows $P_2$ to sparsely retrieve more relevant local responses from $P_1$. Rather than uniformly aggregating all nearby high-resolution signals, it can preserve more useful details for tiny object while reducing the redundant interferences of $P_1$ more effectively. Although AvgPool obtains a slightly higher AP$_s$, deformable cross-attention provides the best overall trade-off between accuracy and complexity, especially on very tiny objects. We therefore use deformable cross-attention as the default downsampling strategy for $P_1$ in CSDF.

\begin{table}[t]
\caption{
Comparison of different strategies for downsampling $P_1$ to the $P_2$ resolution in CSDF. We keep the semantically enhanced $P_1$ and other ReDHF components unchanged, and only vary the strategies. The best result in each column is highlighted in bold.
}
\label{tab:p1_downsampling}
\centering

\setlength{\tabcolsep}{5pt}
\renewcommand{\arraystretch}{1.12}
\begin{tabular}{lccccccccc}
\toprule
Aggregation mode & Params & GFLOPs & AP & AP$_{50}$ & AP$_{75}$ & AP$_{vt}$ & AP$_t$ & AP$_s$ & AP$_m$ \\
\midrule
Bilinear & \textbf{15.88} & \textbf{182.3} & 29.4 & 61.9 & 24.0 & 14.9 & 29.4 & 36.6 & 41.5 \\
AvgPool & \textbf{15.88} & \textbf{182.3} & 33.1 & 66.9 & 28.7 & 18.4 & 32.9 & \textbf{38.8} & 45.0 \\
Strided Conv & 15.95 & 187.5 & 32.8 & 66.7 & 27.9 & 18.1 & 32.8 & 38.0 & 44.7 \\
Deform. Cross-Attn & 15.93 & 188.2 & \textbf{33.5} & \textbf{68.1} & \textbf{28.9} & \textbf{18.9} & \textbf{33.5} & 38.7 & \textbf{45.8} \\
\bottomrule
\end{tabular}
\end{table}

To better understand the large performance gap between different downsampling strategies, we further measure their foreground-to-background response ratios. As shown in Table~\ref{tab:aggregation_fbr}, bilinear interpolation increases FBR by only 0.04 over the original $P_2$, whereas deformable cross-attention increases it by 0.38 and achieves the highest AP$_{vt}$. AvgPool and strided convolution show intermediate improvements. These results indicate that the degradation of bilinear interpolation is associated with indiscriminate mixing of shallow high-resolution responses, while selective aggregation better concentrates the injected information on object regions.

\begin{table}
\caption{
Foreground-to-background response analysis for different downsampling strategies in CSDF.
$\Delta$FBR denotes the increase from $P_2$ to $P_2'$.
The best result in each column is highlighted in bold.
}
\label{tab:aggregation_fbr}
\setlength{\tabcolsep}{4pt}
\renewcommand{\arraystretch}{1.12}
\centering
\begin{tabular}{lcccccc}
\toprule
Aggregation &
$\mu_{\mathrm{in}}(P_2')$ &
$\mu_{\mathrm{out}}(P_2')$ &
FBR$(P_2')$ &
FBR$(P_2)$ &
$\Delta$FBR &
AP$_{vt}$ \\
\midrule
Bilinear
& 0.49 & 0.33 & 1.48 & 1.44 & 0.04 & 14.9 \\
AvgPool
& 0.45 & \textbf{0.27} & 1.67 & 1.45 & 0.22 & 18.4 \\
Strided Conv
& 0.52 & 0.32 & 1.63 & 1.45 & 0.18 & 18.1 \\
Deformable cross-attn
& \textbf{0.57} & 0.31 & \textbf{1.84} & 1.46 &
\textbf{0.38} & \textbf{18.9} \\
\bottomrule
\end{tabular}
\end{table}

\subsection{Sampling points in CSDF and FSQI}
\label{app:sampling_points}
\begin{figure} 
  \centering
  \includegraphics[width=0.9\linewidth]{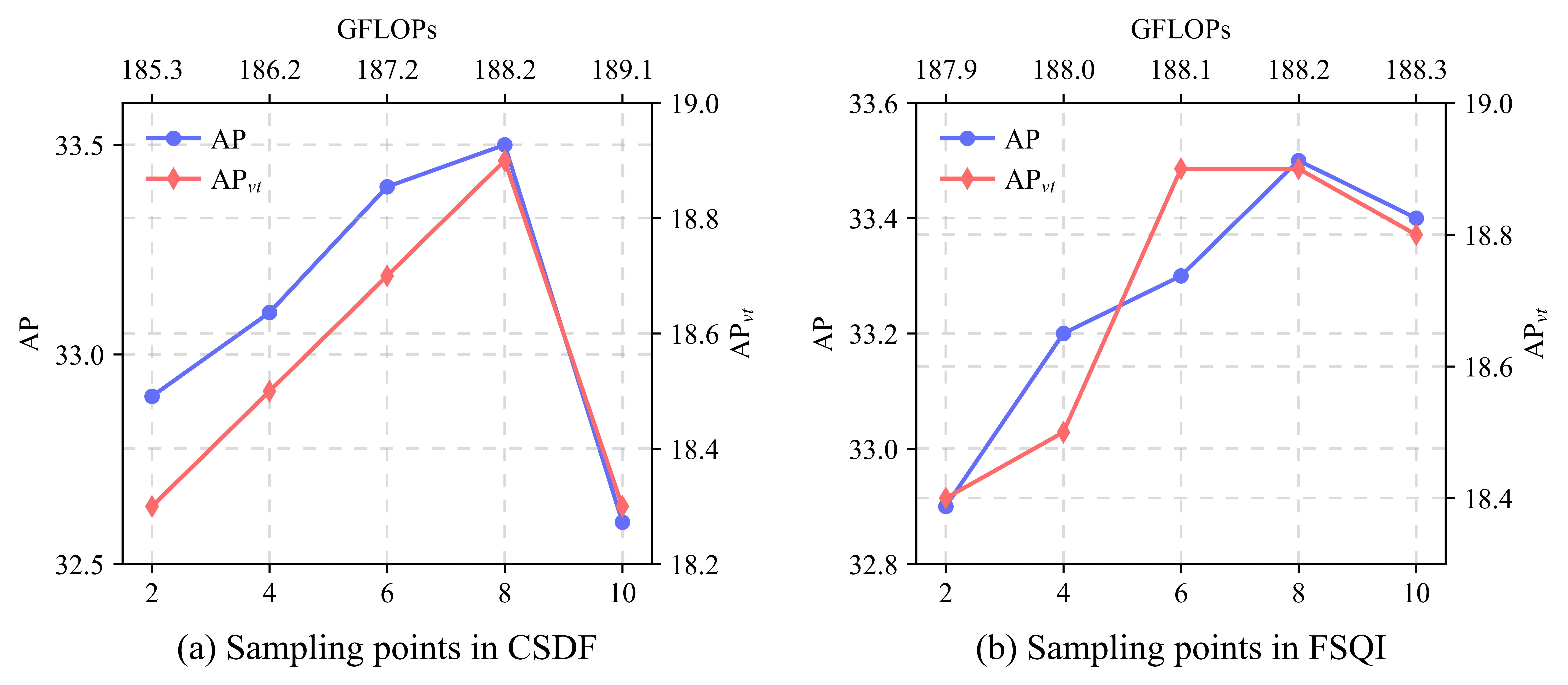}
  
  \caption{Sensitivity analysis of the sampling-point numbers in CSDF and FSQI. The bottom x-axis shows the number of sampling points, the top x-axis shows the corresponding GFLOPs. The left y-axis reports AP, while the right y-axis reports $AP_{vt}$. In the left subplot, we vary the CSDF points while fixing the FSQI points to 8. In the right subplot, we vary the FSQI points while fixing the CSDF points to 8.}

  \label{fig:ablation_points_gflops}
\end{figure}

We further study the influence of sampling-point numbers in CSDF and FSQI, as shown in Fig.~\ref{fig:ablation_points_gflops}. Both modules achieve the best performance with 8 sampling points, suggesting that sufficient sparse sampling is important for capturing fine-grained cues from the highest-resolution feature. However, further increasing the number of sampling points leads to noticeable drops in both AP and AP$_{vt}$, indicating that overly dense sampling may introduce redundant or noisy responses from high-resolution features.

The two modules also show different sensitivities to the sampling-point number. CSDF performs encoder-side detail injection from $P_1$ to $P_2$, so the sampled responses are directly fused into the shallow feature. As a result, inappropriate sampling may more easily introduce background responses and disturb the feature representation. In contrast, FSQI exposes $P_1$ to decoder queries through cross-attention, where each query can adaptively retrieve useful high-resolution responses together with information from other feature levels. This query-driven selection makes FSQI relatively more stable under different sampling settings, although excessive sampling is still harmful. Therefore, we set the number of sampling points to 8 for both CSDF and FSQI by default.

\subsection{Scale threshold in EGSS}
\label{app:rmax}

We further analyze the scale threshold $r_{\max}$ used by EGSS. Since $r_{\max}$ determines which ground-truth objects are considered size-matched to the shallow feature $P_2'$, its appropriate value depends on the object-scale distribution of each dataset.

\begin{table}[t]
\caption{
Sensitivity to the scale threshold $r_{\max}$ in EGSS.
GT coverage denotes the percentage of valid ground-truth instances selected for
supervising $P_2'$ under the corresponding threshold.
Tiny AP denotes AP$_{vt}$, AP$_s$, and AP$_{eS}$ on AI-TOD-v2, VisDrone, and SODA-D,
respectively.
The adopted setting for each dataset is highlighted in bold.
}
\label{tab:rmax_sensitivity}
\setlength{\tabcolsep}{6pt}
\renewcommand{\arraystretch}{1.12}
\centering
{\fontsize{7.0pt}{8.7pt}\selectfont
\begin{tabular}{lccccc}
\toprule
Dataset &
Baseline AP / Tiny AP &
GT coverage @8 / @16 (\%) &
AP / Tiny AP @8 &
AP / Tiny AP @16 &
Default $r_{\max}$ \\
\midrule
AI-TOD-v2 & 32.1 / 17.5 & 98.17 / 100.00 &
\textbf{33.5 / 18.9} & 33.4 / 18.7 & \textbf{8} \\
VisDrone & 35.9 / 28.3 & 64.14 / 90.28 &
37.1 / 29.8 & \textbf{37.5 / 30.3} & \textbf{16} \\
SODA-D & 35.4 / 17.6 & 73.69 / 100.00 &
36.6 / 19.3 & \textbf{36.8 / 19.6} & \textbf{16} \\
\bottomrule
\end{tabular}
}
\end{table}

Table~\ref{tab:rmax_sensitivity} reports both the percentage of valid ground-truth instances covered by each threshold and the resulting detection performance. On AI-TOD-v2, $r_{\max}=8$ already covers 98.17\% of valid ground-truth instances and performs slightly better than $r_{\max}=16$. In contrast, VisDrone and SODA-D contain broader object-scale distributions, for which increasing $r_{\max}$ from 8 to 16 raises the coverage from 64.14\% to 90.28\% and from 73.69\% to 100\%, respectively, with small corresponding performance improvements. Importantly, both settings consistently outperform their baselines on all three datasets. These results show that $r_{\max}$ mainly controls scale coverage rather than requiring sensitive model-specific tuning.

\subsection{Loss design in EGSS}
\label{app:loss_weight}

Table~\ref{tab:egss_loss_design} studies the loss design of EGSS. The classification-only setting already achieves 33.2 AP, suggesting that foreground discrimination is essential for supervising the shallow feature. Adding a lightweight regression loss further improves AP to 33.5 and AP$_{vt}$ from 18.1 to 18.9, indicating that coarse localization guidance is helpful for very tiny objects. However, adding centerness supervision causes a clear drop to 32.1 AP and 15.8 AP$_{vt}$. This suggests that centerness targets are unreliable on noisy shallow features, especially when tiny objects have only a few valid locations.

We further vary the classification-regression weights. A small regression weight works better than a strong one: increasing $\lambda_{\text{reg}}$ from 0.10 to 0.20 or 0.50 reduces AP$_{vt}$ from 18.9 to 17.6 and 17.3, respectively. Reducing the classification weight to 0.5 also degrades performance, showing that foreground classification should remain dominant in EGSS. Therefore, we use $\lambda_{\text{cls}}=1.0$ and $\lambda_{\text{reg}}=0.10$ by default, and omit centerness supervision.

\begin{table}[t]
\caption{
Ablation study on the EGSS loss design. The centerness loss weight is set to 0.1 when used. The best result is highlighted in bold.
}
\label{tab:egss_loss_design}
\centering
\setlength{\tabcolsep}{5.5pt}
\renewcommand{\arraystretch}{1.12}

\begin{tabular}{l c c c c c c c}
\toprule
Setting & $\lambda_{\text{cls}}$ & $\lambda_{\text{reg}}$ & AP & AP$_{50}$ & AP$_{vt}$ & AP$_t$ & AP$_s$ \\
\midrule

cls only        & 1.0 & --   & 33.2 & \textbf{68.2} & 18.1 & 33.2 & 38.5 \\
cls + reg       & 1.0 & 0.10 & \textbf{33.5} & 68.1 & \textbf{18.9} & \textbf{33.5} & 38.7 \\
cls + reg + ctr & 1.0 & 0.10 & 32.1 & 66.5 & 15.8 & 32.2 & 38.2 \\

\midrule

\multirow{4}{*}{cls + reg}
& 1.0 & 0.05 & 33.3 & 67.1 & 18.7 & 33.3 & \textbf{39.2} \\
& 1.0 & 0.20 & 33.1 & 67.2 & 17.6 & 33.7 & 38.6 \\
& 1.0 & 0.50 & 32.8 & 67.7 & 17.3 & 32.3 & 38.6 \\
& 0.5 & 0.10 & 32.6 & 67.0 & 16.0 & 32.4 & 38.2 \\

\bottomrule
\end{tabular}
\end{table}

To examine whether the degradation caused by centerness results from a particular loss weight, we vary $\lambda_{\mathrm{ctr}}$ from 0 to 0.10. As shown in Table~\ref{tab:centerness_weight}, the performance decreases monotonically as the centerness weight increases, with AP$_{vt}$ dropping from 18.9 without centerness to 18.0 and 15.8 at weights 0.05 and 0.10, respectively. We then analyze the sparsity of centerness targets across object scales as shown in Table~\ref{tab:centerness_statistics}. Very tiny objects have only 1.90 positive locations on average, and 93.90\% of them contain at most one location with a centerness target no smaller than 0.5. In comparison, the number of valid locations increases rapidly with object size. This analysis explains why explicit centerness supervision becomes excessively sparse and unstable for very tiny objects, supporting its exclusion from EGSS.


\begin{table}[t]
\centering
\renewcommand{\arraystretch}{1.12}

\begin{minipage}[t]{0.3\textwidth}
\centering
\caption{
Sensitivity to the centerness loss weight in EGSS.
All other settings are kept unchanged.
}
\label{tab:centerness_weight}
\setlength{\tabcolsep}{10pt}
\begin{tabular}{ccc}
\toprule
$\lambda_{\mathrm{ctr}}$ & AP & AP$_{vt}$ \\
\midrule
0    & \textbf{33.5} & \textbf{18.9} \\
0.05 & 32.9 & 18.0 \\
0.10 & 32.1 & 15.8 \\
\bottomrule
\end{tabular}
\end{minipage}
\hfill
\begin{minipage}[t]{0.65\textwidth}
\centering
\caption{
Robustness of the full ReDHF model across three
random seeds on AI-TOD-v2.
}
\vspace{6.3pt}
\label{tab:random_seeds}
\setlength{\tabcolsep}{10pt}
\begin{tabular}{lcc}
\toprule
Run & AP & AP$_{vt}$ \\
\midrule
Original & 33.52 & 18.88 \\
Seed 2   & 33.44 & 18.74 \\
Seed 3   & 33.50 & 18.82 \\
\midrule
Mean $\pm$ Std. &
\textbf{33.49 $\pm$ 0.04} &
\textbf{18.81 $\pm$ 0.07} \\
\bottomrule
\end{tabular}
\end{minipage}
\end{table}

\subsection{Robustness across random seeds}

To examine whether the reported gains depend on a favorable random initialization, we repeat the full ReDHF configuration using three random seeds. As shown in Table~\ref{tab:random_seeds}, ReDHF achieves $33.49\pm0.04$ AP and $18.81\pm0.07$ AP$_{vt}$. The small standard deviations indicate that the improvements of ReDHF are stable across different random seeds.

\begin{table}[h]
\caption{
Statistics of centerness targets across object scales on AI-TOD-v2.
$N_{\mathrm{pos}}$ denotes the number of positive feature locations assigned to each
ground-truth instance, and $N_{c\geq0.5}$ denotes the number of positive locations
whose centerness target is at least 0.5.
}
\label{tab:centerness_statistics}
\setlength{\tabcolsep}{9pt}
\renewcommand{\arraystretch}{1.12}
\centering
\begin{tabular}{lccccc}
\toprule
Scale &
Num. GT &
Avg. $N_{\mathrm{pos}}$ &
$N_{\mathrm{pos}}\leq2$ (\%) &
Avg. $N_{c\geq0.5}$ &
$N_{c\geq0.5}\leq1$ (\%) \\
\midrule
Very tiny & 47317  & 1.90  & 78.26 & 0.54  & 93.90 \\
Tiny      & 276865 & 6.44  & 23.69 & 1.79  & 33.76 \\
Small     & 45612  & 19.19 & 3.18  & 5.21  & 1.57 \\
Medium    & 6782   & 81.05 & 0.13  & 21.59 & 0.93 \\
\bottomrule
\end{tabular}
\end{table}

\newpage
\section*{NeurIPS Paper Checklist}

\begin{enumerate}

\item {\bf Claims}
    \item[] Question: Do the main claims made in the abstract and introduction accurately reflect the paper's contributions and scope?
    \item[] Answer: \answerYes{}{} 
    \item[] Justification:  We provide a detailed and accurate description of the paper’s contributions
and scope in both the abstract and introduction.

\item {\bf Limitations}
    \item[] Question: Does the paper discuss the limitations of the work performed by the authors?
    \item[] Answer: \answerYes{} 
    \item[] Justification: We discuss the limitations of ReDHF in Sec.~\ref{sec:conclusion} and Appendix~\ref{app:limitations}, including additional computational overhead, scale-dependent gains, and localized degradation for large objects in dense scenes.
    

\item {\bf Theory assumptions and proofs}
    \item[] Question: For each theoretical result, does the paper provide the full set of assumptions and a complete (and correct) proof?
    \item[] Answer: \answerNA{} 
    \item[] Justification: The paper does not include theoretical results.

    \item {\bf Experimental result reproducibility}
    \item[] Question: Does the paper fully disclose all the information needed to reproduce the main experimental results of the paper to the extent that it affects the main claims and/or conclusions of the paper (regardless of whether the code and data are provided or not)?
    \item[] Answer: \answerYes{} 
    \item[] Justification: The paper provide the network architectures and hyperparameters required for the
experiments in Sec.~\ref{method} and Sec.~\ref{imple_detail}.

\item {\bf Open access to data and code}
    \item[] Question: Does the paper provide open access to the data and code, with sufficient instructions to faithfully reproduce the main experimental results, as described in supplemental material?
    \item[] Answer: \answerYes{} 
    \item[] Justification: The code are provided in the supplemental material. The data is public available and have been widely used.

\item {\bf Experimental setting/details}
    \item[] Question: Does the paper specify all the training and test details (e.g., data splits, hyperparameters, how they were chosen, type of optimizer) necessary to understand the results?
    \item[] Answer: \answerYes{} 
    \item[] Justification: We provide detailed descriptions of all the training and test details, including
backbones, hyperparameters, and training recipes in Sec.~\ref{imple_detail}.

\item {\bf Experiment statistical significance}
    \item[] Question: Does the paper report error bars suitably and correctly defined or other appropriate information about the statistical significance of the experiments?

    \item[] Answer: \answerYes{}
    \item[] Justification: We report the mean and standard deviation across three random seeds for the full ReDHF model on AI-TOD-v2, as shown in Table~\ref{tab:random_seeds}.
        

\item {\bf Experiments compute resources}
    \item[] Question: For each experiment, does the paper provide sufficient information on the computer resources (type of compute workers, memory, time of execution) needed to reproduce the experiments?
    \item[] Answer: \answerYes{} 
    \item[] Justification: We provide the hardware specifications required for the experiments.
    
\item {\bf Code of ethics}
    \item[] Question: Does the research conducted in the paper conform, in every respect, with the NeurIPS Code of Ethics \url{https://neurips.cc/public/EthicsGuidelines}?
    \item[] Answer: \answerYes{} 
    \item[] Justification: The research conducted in the paper conform with the NeurIPS Code of Ethics

\item {\bf Broader impacts}
    \item[] Question: Does the paper discuss both potential positive societal impacts and negative societal impacts of the work performed?
    \item[] Answer: \answerYes{} 
    \item[] Justification: Our method significantly improves tiny object detection and can benefit applications such as remote sensing, traffic monitoring, and safety inspection.
    
\item {\bf Safeguards}
    \item[] Question: Does the paper describe safeguards that have been put in place for responsible release of data or models that have a high risk for misuse (e.g., pre-trained language models, image generators, or scraped datasets)?
    \item[] Answer: \answerYes{} 
    \item[] Justification: Our method exclusively utilizes open-source models and datasets, ensuring that there is no risk of misuse.

\item {\bf Licenses for existing assets}
    \item[] Question: Are the creators or original owners of assets (e.g., code, data, models), used in the paper, properly credited and are the license and terms of use explicitly mentioned and properly respected?
    \item[] Answer: \answerYes{} 
    \item[] Justification: Our method exclusively utilizes open-source models and datasets.

\item {\bf New assets}
    \item[] Question: Are new assets introduced in the paper well documented and is the documentation provided alongside the assets?
    \item[] Answer: \answerNA{} 
    \item[] Justification: The paper does not release new assets.

\item {\bf Crowdsourcing and research with human subjects}
    \item[] Question: For crowdsourcing experiments and research with human subjects, does the paper include the full text of instructions given to participants and screenshots, if applicable, as well as details about compensation (if any)? 
    \item[] Answer: \answerNA{} 
    \item[] Justification: This study does not involve any crowdsourced experiments or research with human subjects.

\item {\bf Institutional review board (IRB) approvals or equivalent for research with human subjects}
    \item[] Question: Does the paper describe potential risks incurred by study participants, whether such risks were disclosed to the subjects, and whether Institutional Review Board (IRB) approvals (or an equivalent approval/review based on the requirements of your country or institution) were obtained?
    \item[] Answer: \answerNA{} 
    \item[] Justification: The paper does not involve crowdsourcing nor research with human subjects.

\item {\bf Declaration of LLM usage}
    \item[] Question: Does the paper describe the usage of LLMs if it is an important, original, or non-standard component of the core methods in this research? Note that if the LLM is used only for writing, editing, or formatting purposes and does \emph{not} impact the core methodology, scientific rigor, or originality of the research, declaration is not required.
    \item[] Answer: \answerNA{} 
    \item[] Justification: The core method development in this research does not involve LLMs as any important, original, or non-standard components.

\end{enumerate}

\end{document}